\documentclass[Afour,sageh,times]{sagej}

\usepackage{moreverb,url}

\newcommand{\thh}{\textsuperscript{th}}

\newcommand{\rd}{\textsuperscript{rd}}

\usepackage{subfigure}

\usepackage[colorlinks,bookmarksopen,bookmarksnumbered,citecolor=red,urlcolor=red]{hyperref}

\newcommand\BibTeX{{\rmfamily B\kern-.05em \textsc{i\kern-.025em b}\kern-.08em
T\kern-.1667em\lower.7ex\hbox{E}\kern-.125emX}}

\def\volumeyear{2026}

\newenvironment{contributions}[1]%
{\subsection*{\normalsize\sagesf\bfseries Author contributions}\begin{refsize}\noindent #1}%
{\end{refsize}}

\begin{document}

\runninghead{Fiorista, et al.}

\title{Closed-Circuit Television Data as an Emergent Data Source in Urban Rail Platform Crowding Estimation}

\author{
Riccardo Fiorista\affilnum{1},
Awad Abdelhalim\affilnum{1},
Anson Stewart\affilnum{1},
Gabriel L. Pincus\affilnum{2},
Ian Thistle\affilnum{2} and
Jinhua Zhao\affilnum{1}
}

\affiliation{
\affilnum{1}Transit Lab, Department of Urban Studies and Planning,
Massachusetts Institute of Technology, Cambridge, MA 02139, USA\\
\affilnum{2}Washington Metropolitan Area Transit Authority,
300 7th Street SW, Washington, DC 20024, USA
}

\corrauth{
Riccardo Fiorista,
Transit Lab,
Department of Urban Studies and Planning,
Massachusetts Institute of Technology,
Cambridge, MA 02139, USA.
}

\email{riccardo.fiorista@mit.edu}

\begin{abstract}
Accurately estimating urban rail platform occupancy can support more informed operational decision-making by transit agencies, particularly during periods of crowding. However, sensing real-time platform occupancy remains challenging and often relies on indirect proxies, such as automatic fare collection data or staff observations. Recently, Closed-Circuit Television (CCTV) footage has emerged as a promising data source for accurate, real-time occupancy estimation. The present study investigates this approach by comparing three state-of-the-art computer vision approaches for extracting crowd-related features from platform CCTV imagery: (a) object detection and counting using YOLOv11, RT-DETRv2, and APGCC; (b) crowd-level classification via a custom-trained Vision Transformer (Crowd-ViT); and (c) semantic segmentation using DeepLabV3. Additionally, we present a novel and efficient convex ridge regression over mean-pooled segmentation features to extract counts from the generated segmentation maps while accounting for image depth and passenger dispersion along a platform. Tested on a privacy-preserving dataset, created in collaboration with the Washington Metropolitan Area Transit Authority (WMATA), and including more than 600 hours of video material, our results demonstrate that computer vision approaches can provide substantive value for crowd estimation. This work demonstrates that CCTV image data, independent of other data sources available to transit agencies, can enable real-time crowding estimation and, eventually, timely operational responses to mitigate platform crowding.
\end{abstract}

\keywords{Computer Vision, Urban Rail, Crowding}

\maketitle

\section{Introduction \& Related Work}
Accurately estimating rail ridership during revenue service is essential for safe, efficient, and customer-focused urban rail operations \cite{miller_estimation_2018, tirachini_crowding_2013}. Most prior work focuses on in-vehicle crowding \cite{li_crowding_2011}, whereas platform-level crowding warrants separate attention given its implications for passenger safety, operational efficiency, and passenger experience. Existing methods, including dynamic factor models \cite{noursalehi_predictive_2021}, regression trees \cite{kopsidas_extracting_2023}, and graph neural networks \cite{10462016}, often focus on short-term prediction by reconstructing occupancy from historical data, Automatic Fare Collection (AFC) records, and train movement data. Although these data sources are typically available in real time, ground-truth platform occupancy is not, limiting the effectiveness of such estimation methods, particularly for autoregressive time-series models. In this context, Closed-Circuit Television (CCTV) footage has the potential to provide such occupancy estimates in real-time, increasing operational awareness and aiding short-term prediction.

To extract information from such CCTV imagery, Computer Vision (CV) techniques are needed. Recent CV advancements offer substantial gains in automated object detection and scene understanding, improving accuracy and inference speed \cite{saleh_recent_2015, lecun_convolutional_2010}. Notably, neural network-based architectures such as You Only Look Once (YOLO) \cite{Redmon_2016_CVPR} and Detection Transformer (DETR) \cite{Zhao_2024_CVPR} have already been successfully employed for people counting in public transit \cite{florez_monitoring_2023, goh_image_2018}. Domain-specific methods, including Auxiliary Point Guidance Crowd Counting (APGCC) \cite{chen_improving_2024, cheng_rethinking_2022} and density map estimation \cite{khan_advances_2020}, have further shown improvement over general methods for crowd counting accuracy in densely populated scenes, where traditional detection methods face limitations due to occlusions and image artifacts. Model-agnostic techniques, such as Slicing Aided Hyper Inference (SAHI) \cite{akyon_slicing_2022}, which performs inference on smaller image patches, have been developed to further address such task-specific challenges.

Beyond detection, recent research has demonstrated the effectiveness of image classification and semantic segmentation in various contexts. Vision Transformers (ViT) \cite{dosovitskiy_image_2021} have shown excellent classification performance in transportation-related tasks, such as predicting bus delays from traffic camera imagery \cite{abdelhalim_computer_2024}. Additionally, semantic segmentation approaches such as DeepLabV3 \cite{chen_deeplab_2017} and U-Net \cite{ronneberger2015u} provide pixel-level scene understanding, further enriching the set of available CV tools for identifying image regions containing specific objects.

Despite these advancements, the potential of CV-derived features for real-time platform crowding estimation remains underexplored. Existing research has primarily focused on in-station passenger flow \cite{wong_fusion_2023} or vehicle counts \cite{peppa_urban_2018}, with limited investigation into continuous, real-time CCTV-based occupancy estimation. A notable exception is the study in \cite{wang_2025_assessment}, which applied a customized YOLOv8 model with additional attention layers to detect individuals over a single day's on-platform CCTV footage at a Beijing metro station. Their work yielded promising, yet noisy initial results on passenger counts.

This paper comprehensively evaluates the use of CCTV-derived features to estimate rail platform occupancy. We use a privacy-preserving, self-collected, labeled CCTV dataset covering multiple platforms and revenue service days within the Washington Metropolitan Area Transit Authority (WMATA) Metrorail network. Specifically, we investigate three CV approaches:

\begin{enumerate}
    \item \textbf{Crowd counting:} YOLOv11 \cite{jocher_ultralytics_2023} and the Real Time (RT)-DETRv2 \cite{lv_rt-detrv2_2024} object detection, both enhanced with Slicing-Aided Hyper Inference (SAHI), as well as APGCC \cite{chen_improving_2024} head counting;
    \item \textbf{Crowd-level classification:} Custom fine-tuned ViT model, which we refer to as \textit{Crowd-ViT} \cite{dosovitskiy_image_2021};
    \item \textbf{Segmentation-based crowd estimation:} DeepLabV3 model \cite{chen_rethinking_2017} with a novel camera-specific spatial Ridge Regression (RR)-based calibration approach.
\end{enumerate}

We evaluate the performance of these image-based approaches against the derived reference occupancy obtained from WMATA's internal implementation of the Origin-Destination Interchange (ODX) algorithm \cite{gordon_estimation_2018}. Specifically, we assess model accuracy at three aggregation levels: individual images, train-arrival events, and 15-minute mean aggregates.

\section{Methodology}

This section introduces the ODX-derived platform occupancy estimates used as our reference occupancy for image labeling, followed by a description of the generated CCTV dataset and the CV methods used to estimate platform crowding: object detection, classification, and segmentation. We then present our RR-based calibration method that enables direct occupancy estimation from segmentation maps. Since ODX-derived labels inherit uncertainty from the underlying passenger-assignment process, we quantify the associated measurement noise using a state-space model (SSM), thereby contextualizing the accuracy of the proposed CV-based estimates. We conclude with an overview of the metrics used to quantify performance.

\begin{figure}[h]
    \centering
    \includegraphics[width=\linewidth]{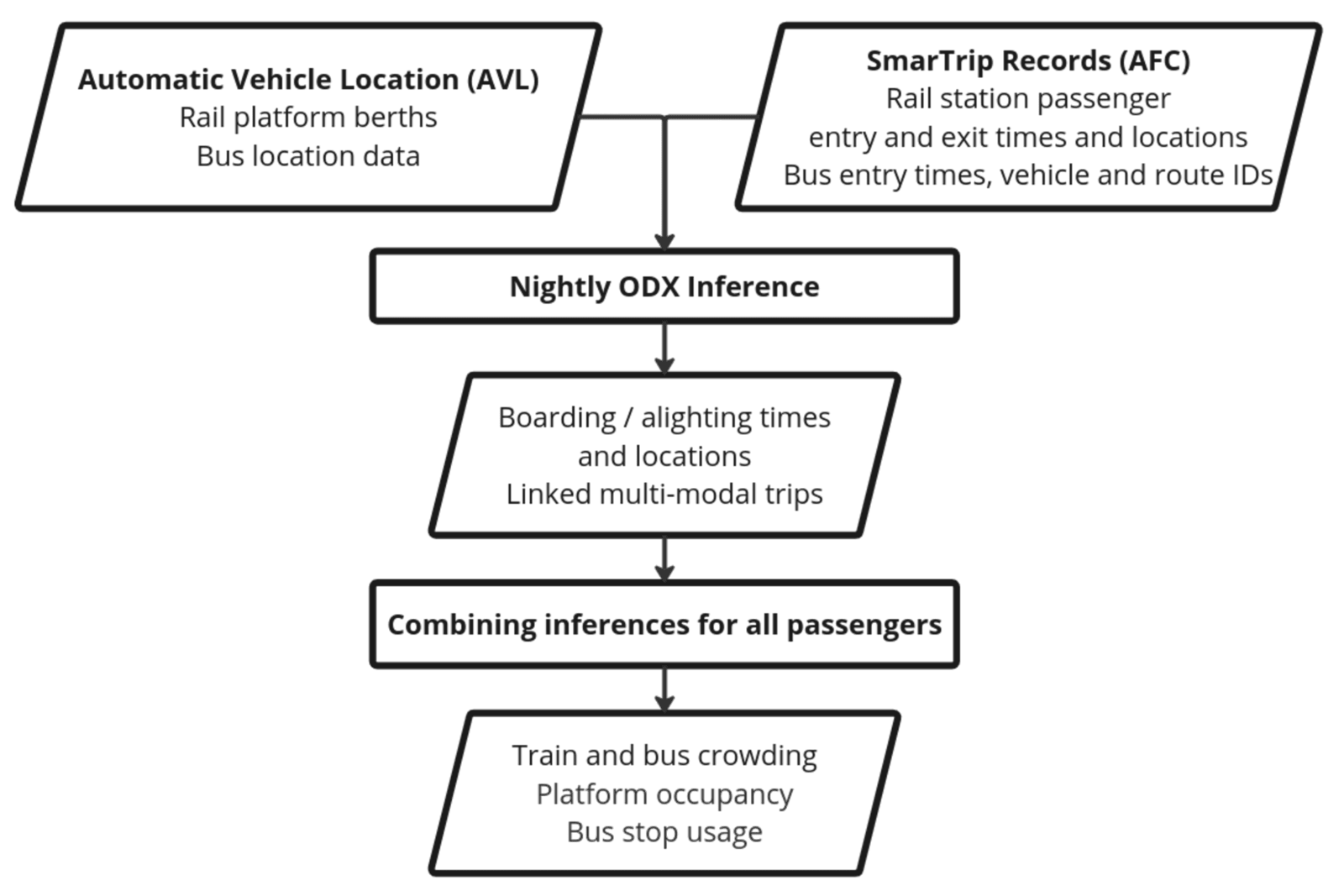}
    \caption{High-level Origin-Destination Interchange (ODX) algorithm diagram. Smartcard data from WMATA's closed-loop Automatic Fare Collection (AFC) system is combined with Automatic Vehicle Location (AVL) data. After the end of a revenue service day, ODX reconstructs each rider's journey and provides probability-based estimates of rail platform and bus berth occupancies, vehicle loads, and travel times.}
    \label{ch2:fig:odx-explanation}
\end{figure}

\subsection{Reference Occupancy Data}
\label{ch3:met:ground_truth}

The reference rail platform occupancy values in this study are derived from WMATA's ODX algorithm \cite{gordon_estimation_2018}, which combines AFC smart card tap data, Automatic Vehicle Location (AVL) data, and operational context to infer likely rider trajectories. Supporting both tap-in (semi-closed) and tap-in/out (closed-loop) systems, ODX assigns probabilities across feasible trajectories to estimate platform occupancies throughout the network, as illustrated in \autoref{ch2:fig:odx-explanation}. These values should therefore be interpreted as probabilistic operational reference estimates rather than direct observations of platform crowding. Uncertainty is greatest at transfer stations, where fare transactions do not capture metro-service transfers and interlined train movements can make multiple rider trajectories plausible. The resulting occupancy estimate thus represents an expectation under the ODX assignment model. We quantify this reference-label uncertainty in the \textit{ODX Error Assessment} section using the state-space decomposition described below.

\begin{figure}
    \centering
    \includegraphics[width=\linewidth]{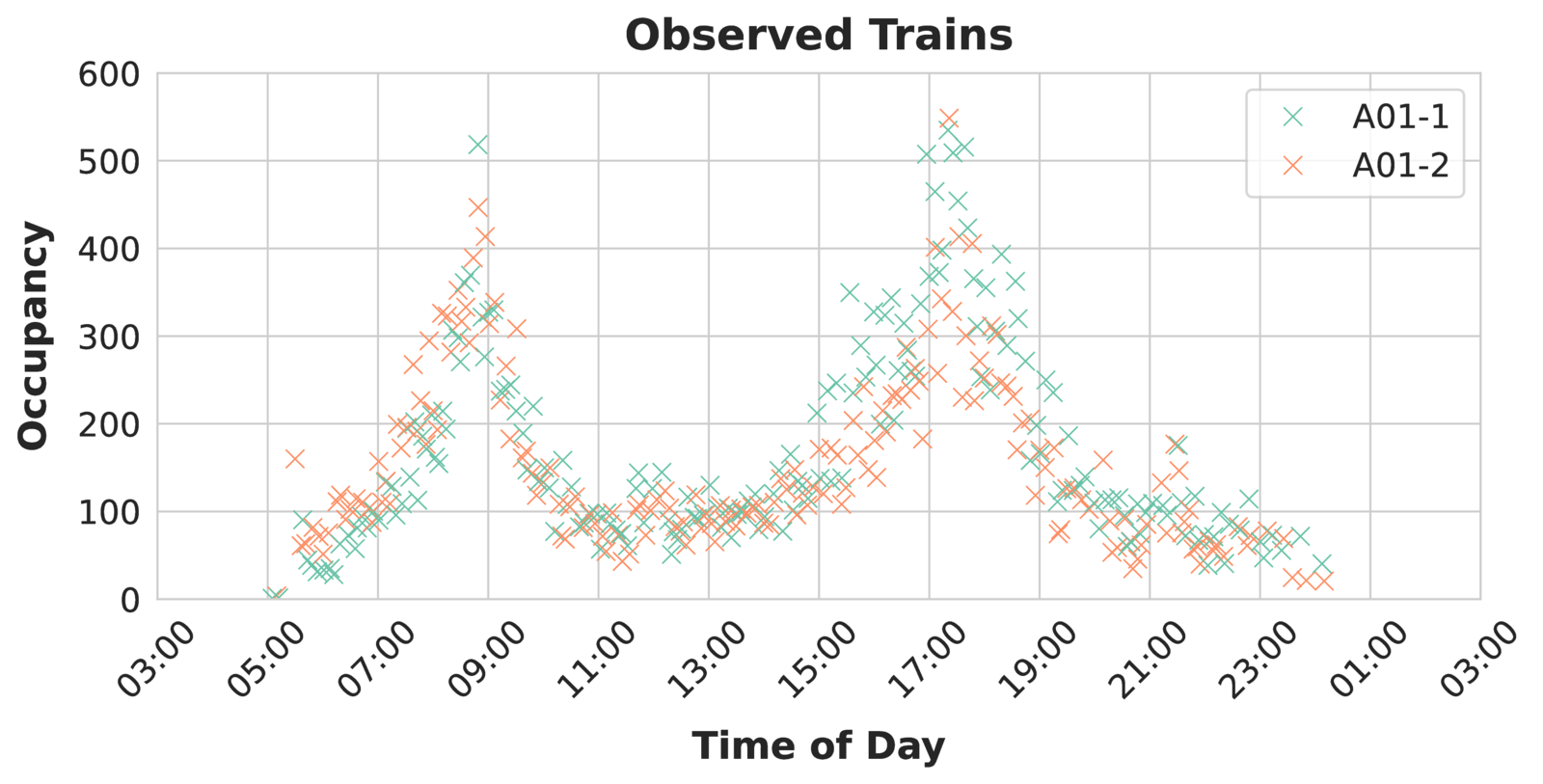}
    \caption{Sample of observed trains over a revenue service day on Metro Center platforms A01-1/2 with ODX-estimated platform occupancies on the y-axis attributed to the individual train-arrival events over time.}
    \label{fig:observed_trains}
\end{figure}

ODX produces continuous-valued occupancy estimates, $y \in \mathbb{R}^+$, for each platform-level train-arrival event (see \autoref{fig:observed_trains}). Each estimate corresponds to the moment of peak platform crowding, internally referred to as the \textit{magic moment}, when a train has arrived, all passengers have alighted, and none have yet boarded. Because ODX computes these estimates retrospectively overnight, neither real-time platform occupancy estimates nor reconstructed individual rider trajectories are available during revenue service.

We use these per-train-arrival event estimates to label CCTV images and compute contiguous 15-minute mean occupancies. For CCTV-based occupancy estimation, we do not leverage any other exogenous variables beyond the image data from on-platform CCTV cameras.

\subsection{CCTV Data}
\label{ch3:met:data_prep}

This section details our video collection and image sampling approach. We illustrate how ODX estimates are attributed to images for counting-based approaches (object detection, head counting, and segmentation), and how we establish crowd-level labels for the Crowd-ViT classification approach.

\subsubsection{Data Acquisition and Sampling}
\label{ch3:met:cctvacquisition}

\begin{table}[!h]
    \centering
    \caption{Summary of available CCTV data for Metro Center (A01/C01), 
    DuPont Circle (A03), Gallery Place-Chinatown (B01), and Navy Yard-Ballpark 
    (F05). Side platforms are identified by \textit{1/2}; center platforms by 
    \textit{C}. Counts are reported by crowd level (Very Light, Light, Medium, and High), depending on observed occupancy percentiles (see the \textit{Crowd-Level Generation} section).}
    \label{tab:data_per_platform}
    \resizebox{\linewidth}{!}{%
        \begin{tabular}{lllllllll}
        \toprule
        \textbf{Platform} & \textbf{\#Cameras} & \textbf{\#Images} & \textbf{\#Train-Arrival Events} & \textbf{\#All Captured} & \textbf{\#VERY LIGHT} & \textbf{\#LIGHT} & \textbf{\#MEDIUM} & \textbf{\#HIGH} \\
        \midrule
        A01-1 & 2 & 4,310 & 567 & 408 & 103 & 136 & 288 & 40 \\
        A01-2 & 3 & 4,041 & 517 & 0   & 113 & 135 & 234 & 35 \\
        A03-1 & 3 & 1,036 & 80  & 80  & 16  & 25  & 36  & 3 \\
        A03-2 & 1 & 352   & 80  & 80  & 18  & 26  & 33  & 3 \\
        B01-2 & 1 & 88 & 20 & 20 & 5 & 2 & 6 & 7 \\
        C01-C & 6 & 33,253 & 1,285 & 1,176 & 338 & 229 & 596 & 122 \\
        F05-C & 2 & 152 & 17 & 17 & 0 & 0 & 1 & 16 \\
        \midrule
        \textbf{Totals} & \textbf{18} & \textbf{43,232} & \textbf{2,566} & \textbf{1,781} & \textbf{593} & \textbf{553} & \textbf{1,194} & \textbf{226} \\
        \bottomrule
        \end{tabular}
    }
\end{table}

The video dataset was collected in collaboration with the WMATA Data Lab. WMATA's rail network comprises 123 platforms across 97 stations, monitored by over 1,000 cameras at fare gates, on platforms, and along rail tracks. This study focuses on seven platforms across four central Washington D.C. stations, namely, Metro Center, DuPont Circle, Gallery Place-Chinatown, and Navy Yard, using footage from 18 platform-mounted cameras. Videos were sampled, following strict internal privacy guidelines, between June 3\rd, 2024, and January 20\thh, 2025, during weekday peak hours (07:00-09:00 and 16:00-19:00) and high-demand events such as July 4\thh$\;$and major sporting events. The selected stations and corresponding platform-camera configurations, with the platforms covered by our camera data in bold, are:

\begin{itemize}
    \item \textit{Metro Center}: \textbf{A01-1/2} (Red Line, side platforms); \textbf{C01-C} (Blue/Orange/Silver, center platform).
    \item \textit{Gallery Place-Chinatown}: \textbf{B01-}1/\textbf{2} (Red Line, one of two side platforms); F01-C (Green/Yellow, center platform).
    \item \textit{Dupont Circle}: \textbf{A03-1/2} (Red Line, side platforms).
    \item \textit{Navy Yard-Ballpark}: \textbf{F05-C} (Green Line, center platform).
\end{itemize}

\begin{figure}[htbp]
    \centering
    \includegraphics[width=0.32\linewidth]{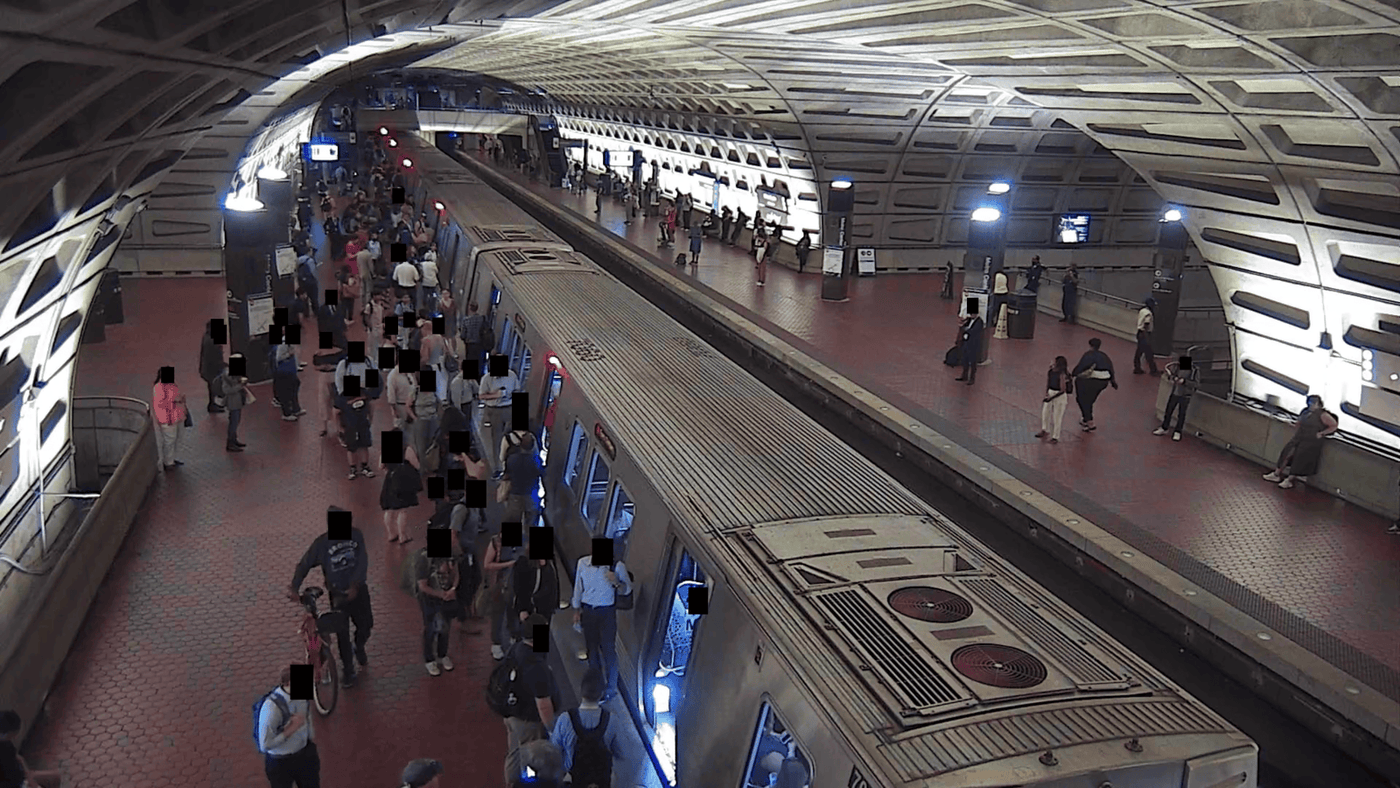}
    \includegraphics[width=0.32\linewidth]{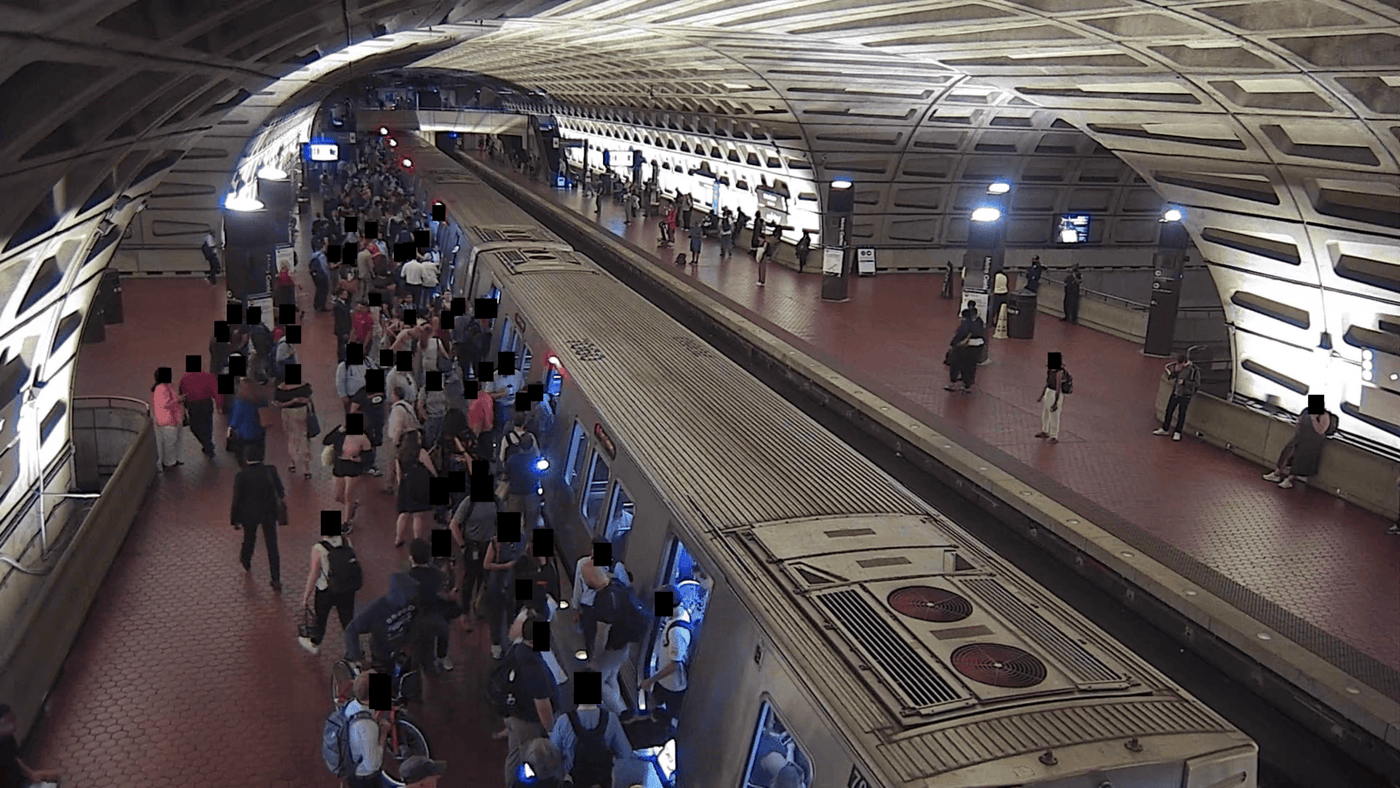}
    \includegraphics[width=0.32\linewidth]{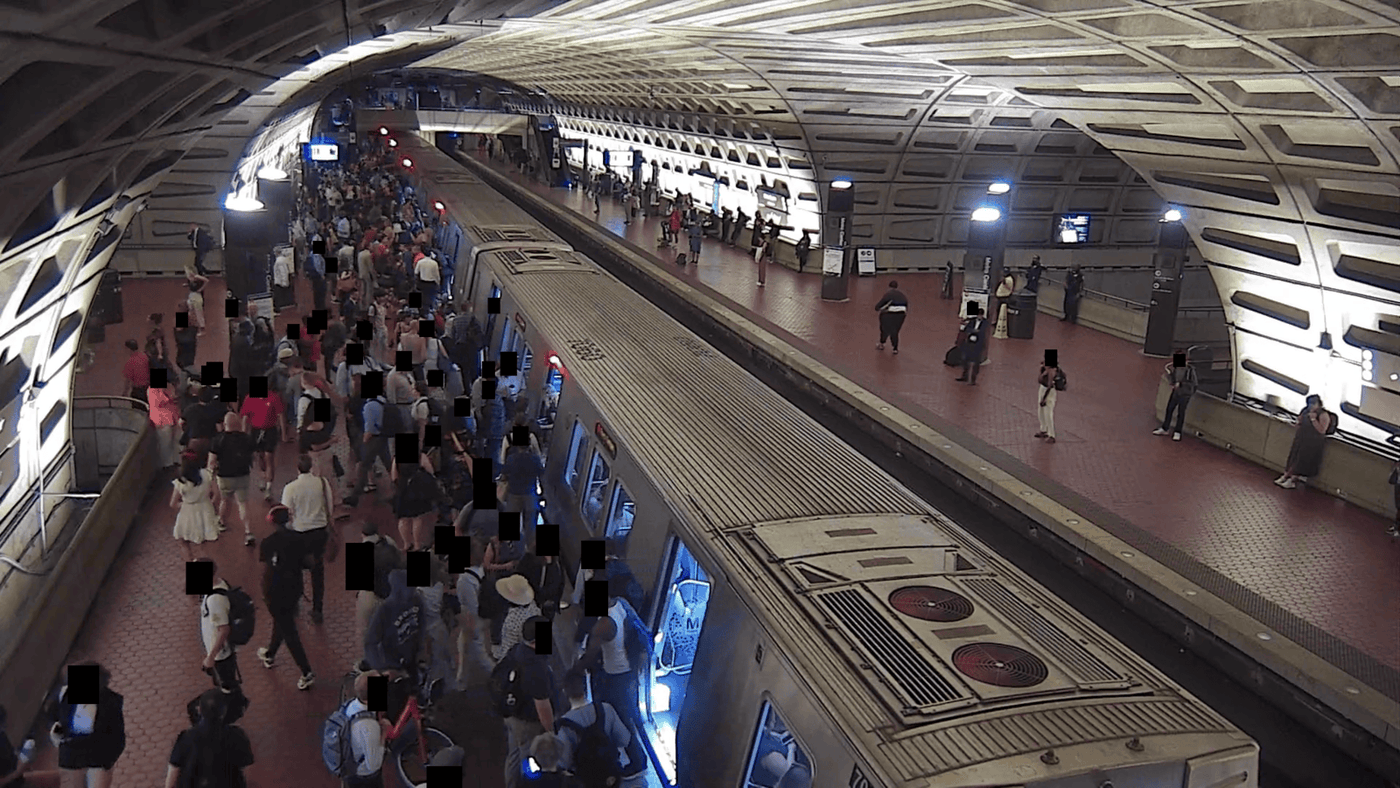}
    \caption{Three sample frames captured during a train-arrival event from a camera at Metro Center platform A01-2 showing the progression around the magic moment (-5s, 0s, +5s) of a scene at the \textit{high} crowd level. Note how +5s features a visually more crowded scene, highlighting the need for temporal variation in CCTV sampling.}
    \label{fig:progression-samples}
\end{figure}

The sampled footage was then processed to extract three frames per train-arrival event: 5 seconds before, at, and 5 seconds after the previously introduced magic moment. The ODX occupancy estimates collected for each train-arrival event serve as a noisy proxy for the ground-truth platform occupancies. This approach mitigates timing misalignment and increases visual diversity, improving dataset robustness. \autoref{fig:progression-samples} shows an example sequence in high crowding, and \autoref{tab:data_per_platform} summarizes the dataset. We note that the resulting image dataset is imbalanced due to variation in sampling periods and the natural, long-tailed distribution of platform crowding (see \textit{Crowd-Level Classification Data Augmentation}). With most samples, Metro Center receives particular emphasis in our analysis. For Metro Center platform A01-2, no train-arrival event contains valid imagery from all sampled cameras simultaneously, which explains the zero in the \textit{All Captured} column of \autoref{tab:data_per_platform}.

\subsubsection{Image Pre-Processing}
\label{ch3:met:data_prep:images}

\begin{figure}[htbp]
    \centering
    \includegraphics[width=0.8\linewidth]{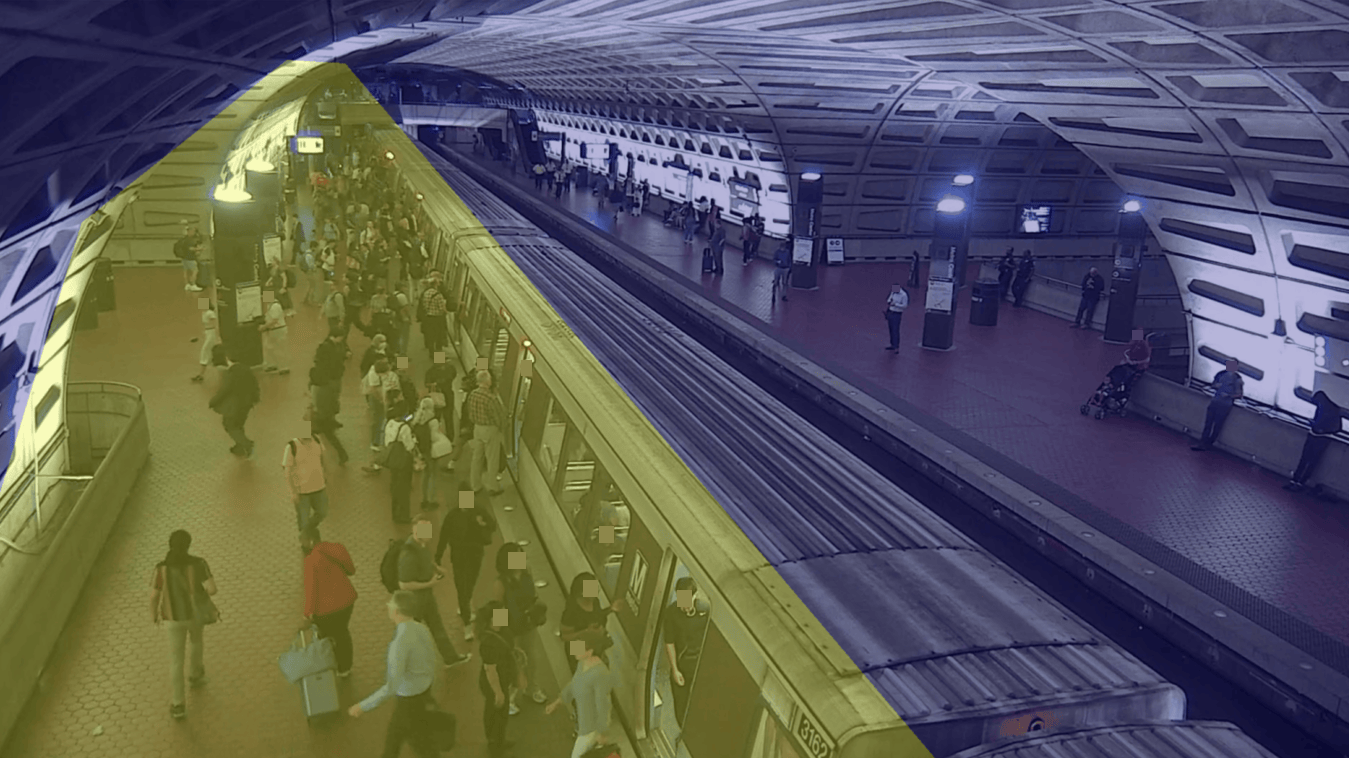}
    \caption{Sample footage of the Metro Center Red line (A01-1/2) platforms with an overlaid area-of-interest mask indicating the region to which CV-based results for one of the platforms are limited to.}
    \label{fig:mask-overlay}
\end{figure}

Despite suboptimal lighting and image quality, each camera has a fixed image resolution, field of view, and viewing angle throughout the dataset. These consistent camera-specific conditions enable effective pre-processing. Thus, we can compute and apply per-camera color normalization values to the respective images. The static views further allow for masking specific \textit{areas of interest}, as illustrated in \autoref{fig:mask-overlay}, which we use in post-processing to prevent cross-platform inference when multiple platforms appear in a single frame.

\subsubsection{CCTV Privacy Considerations}
\label{ch3:met:data_prep:privacy}

All computer vision tasks in this study operate strictly on anonymized, non-identifiable imagery. None of the applied models contains or uses facial recognition components, nor do they attempt to identify, track, or profile individuals. The CV methods extract aggregate counts, scene-level features (such as pixel activations), object-class bounding boxes, or keypoints. All downstream analyses are then performed at the platform level without reference to individual riders. WMATA reviewed these procedures to ensure alignment with internal privacy requirements.

\subsection{Crowd Counting}
\label{ch3:met:crowdcounting}

\subsubsection{Object Detection}
\label{ch3:met:object_detection}

\begin{figure}[!ht]
    \centering

    \subfigure[Original CCTV footage with area-of-interest masking.
    \label{fig:detection:original}]{
        \includegraphics[width=0.3\linewidth]{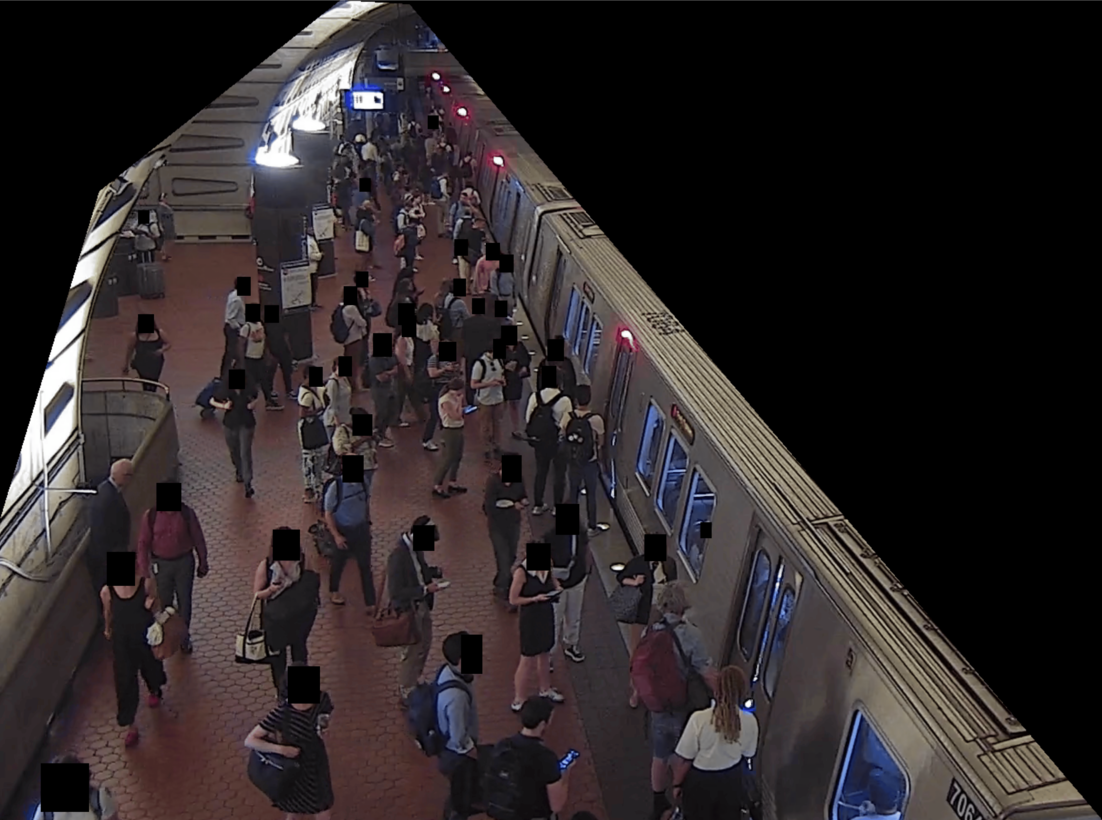}
    }
    \hfill
    \subfigure[YOLOv11 with SAHI, limited to class \textit{Person}.
    \label{fig:detection:YOLOv11}]{
        \includegraphics[width=0.3\linewidth]{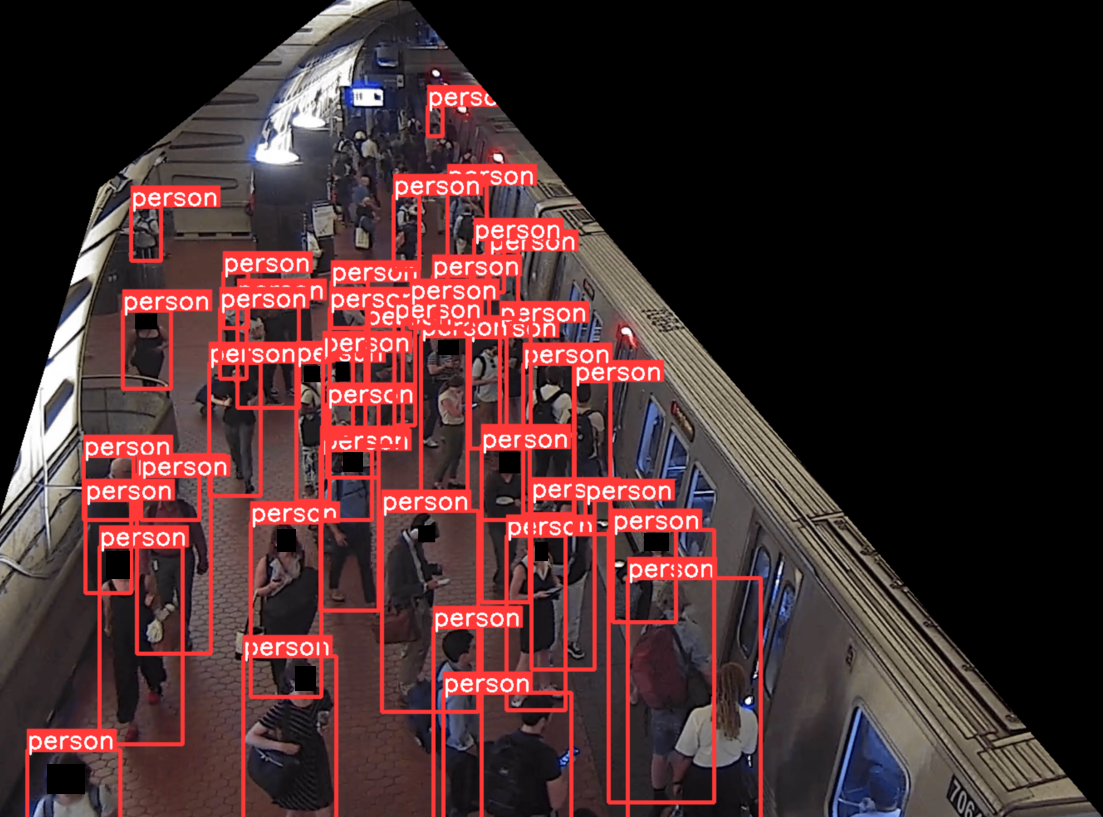}
    }
    \hfill
    \subfigure[RT-DETRv2 with SAHI, limited to class \textit{Person}.
    \label{fig:detection:rtdetrv2}]{
        \includegraphics[width=0.3\linewidth]{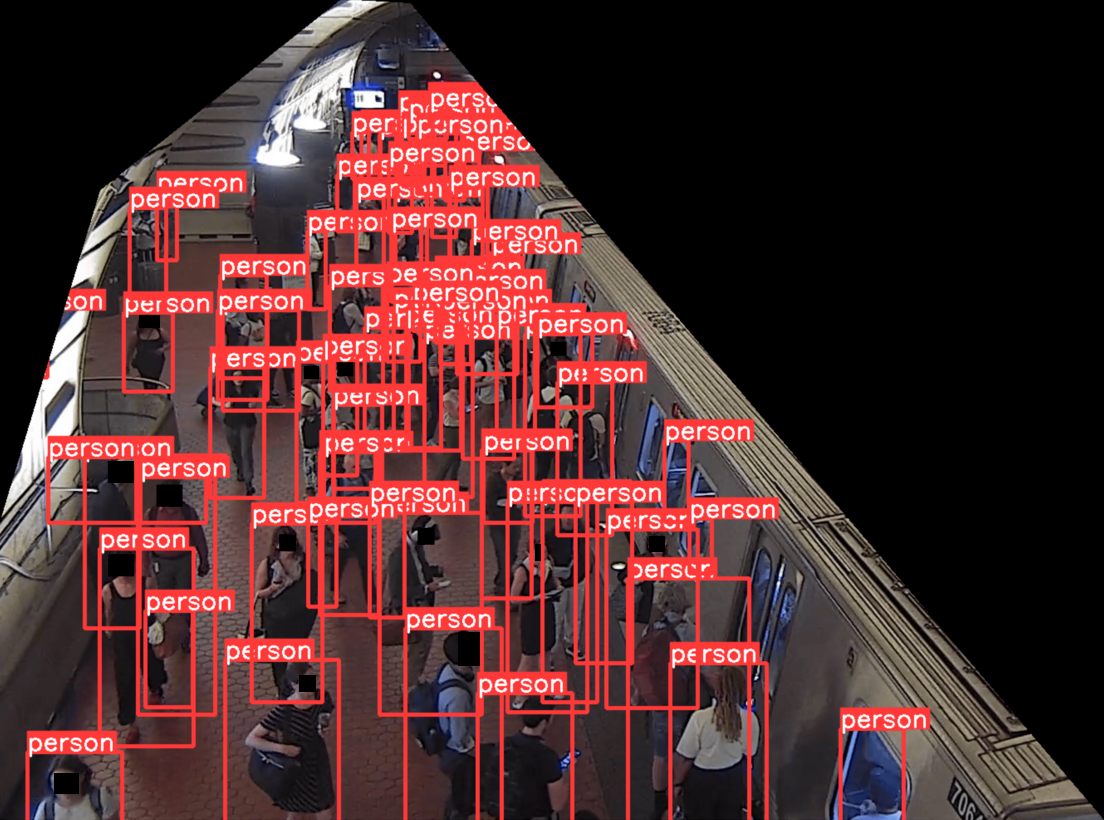}
    }

    \caption{Detection results on Metro Center platform A01-1 from YOLOv11 and RT-DETRv2 (with SAHI), compared to the original masked CCTV frame. Detections use a 30\% confidence threshold, and area-of-interest masking is applied.}
    \label{fig:detection}
\end{figure}

\begin{figure}[!ht]
    \centering

    \subfigure[RT-DETRv2 estimates without SAHI, showing detections of COCO class 0 (person).
    \label{fig:sahicomparison:rtdetrv2wo}]{
        \includegraphics[width=0.45\linewidth]{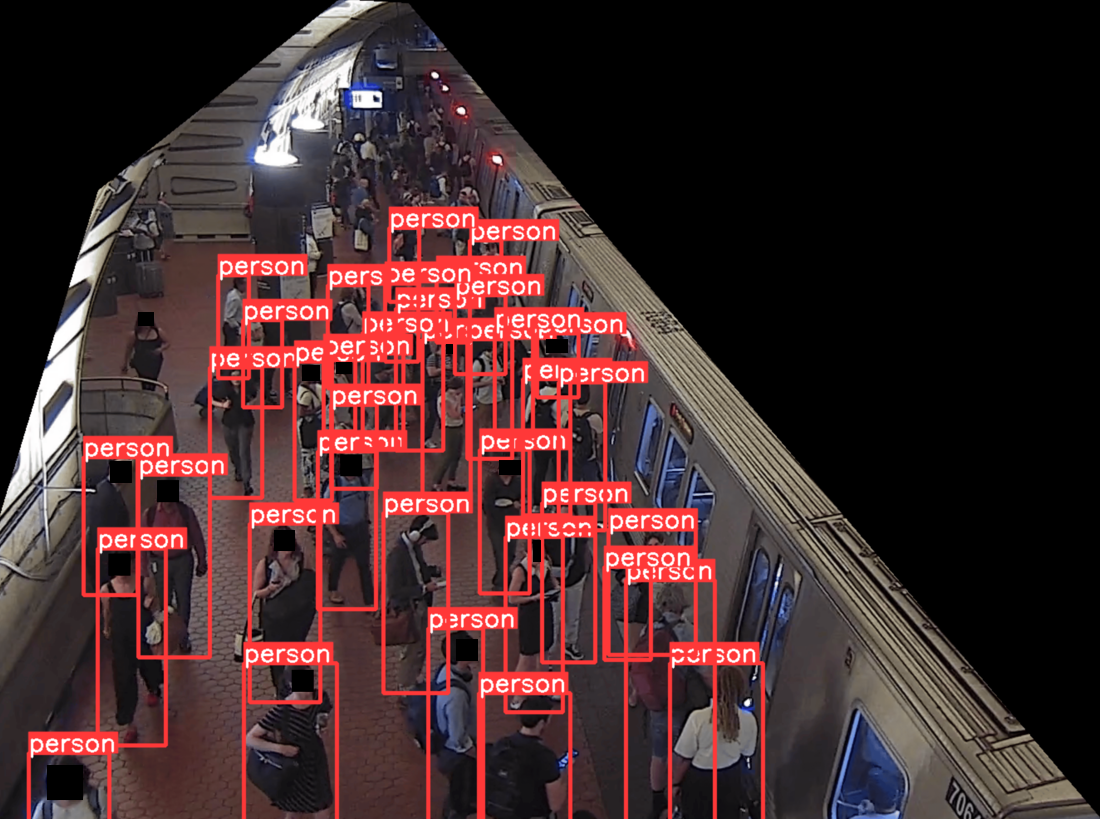}
    }
    \hfill
    \subfigure[RT-DETRv2 with SAHI, applied to the same scene, for COCO class 0 (person).
    \label{fig:sahicomparison:rtdetrv2w}]{
        \includegraphics[width=0.45\linewidth]{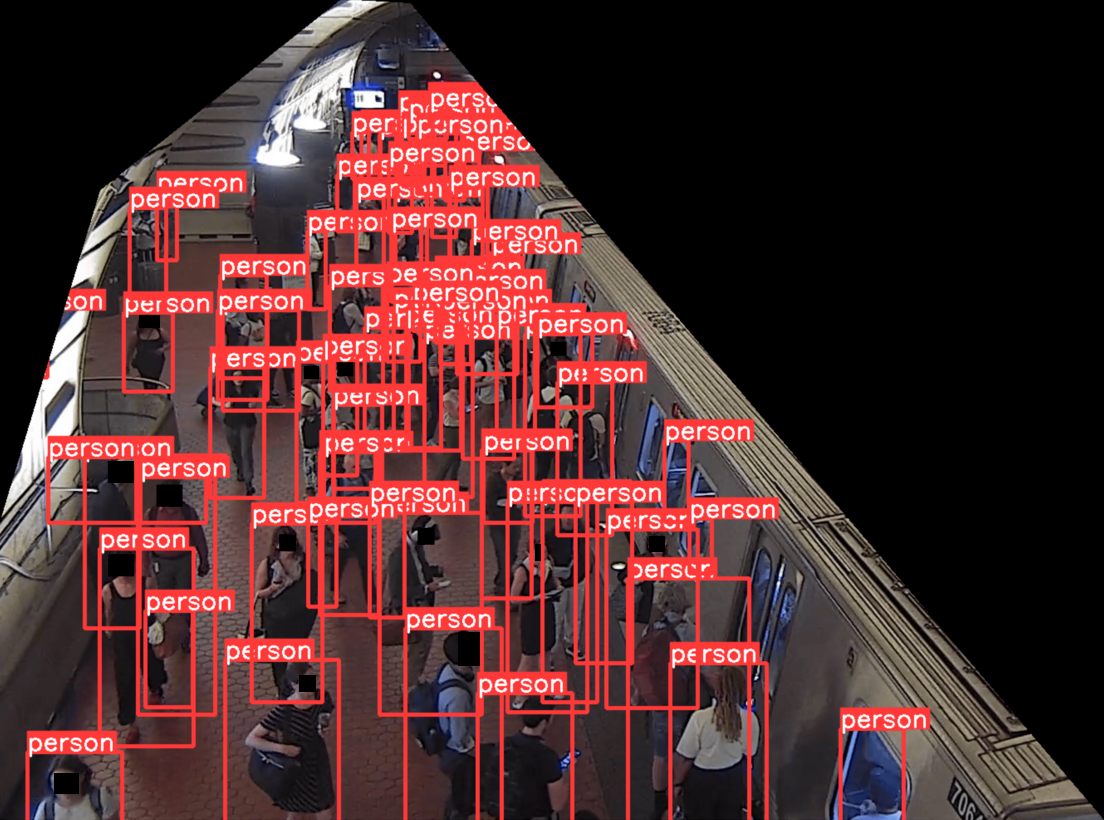}
    }

    \caption{Comparison of RT-DETRv2 detection results on Metro Center platform A01-1. Detections are considered above a 30\% threshold confidence threshold, and if within the area-of-interest mask. Note how, on the right (RT-DETRv2 \textit{with} SAHI), riders in more distant and distorted regions of the image are detected more reliably.}
    \label{fig:sahicomparison}
\end{figure}

Our first approach aims at counting individuals on CCTV images by applying two state-of-the-art object detection models: YOLOv11 (model \textit{YOLOv11x}, 56.9M parameters) \cite{ultralytics_repo} and RT-DETRv2 (model \textit{r101vd}, 76M parameters) \cite{pekingu_rtdetr}, both pre-trained on the Common Objects in Context (COCO) 2017 dataset \cite{lin_microsoft_2014}. Due to variability in image quality across cameras, whose fixed native resolutions range from 500×500 to 1920×1080 pixels, alongside poor lighting, streaming artifacts, and image distortions, we employ the SAHI technique. SAHI processes fixed-size sub-regions (patches) of each image independently, thus improving a model's capability to detect objects in noisy, occluded, and spatially complex images. We illustrate the effectiveness of SAHI when combined with RT-DETRv2 in \autoref{fig:sahicomparison}.

We impose a confidence threshold $\geq30\%$ for both models to minimize false positive detections, which was assessed empirically using randomly sampled images from the training dataset and by evaluating precision vs. recall. Occupancy at the platform level is then estimated by counting the detection bounding boxes for COCO \textit{class 0} (\textit{person}). To reduce temporal noise, we record the maximum person count across the three frames captured (-5s, 0s, +5s) around the magic moment of each train-arrival event. We discuss both results on a per-image, per-train-arrival event, and per-15-minute-bin basis. We compute the maximum occupancy count across the three images for the per-train-arrival event approach. For the 15-minute bin aggregation, we instead consider the \textit{mean} of all maximum values sampled in the respective bin.

\subsubsection{Head Counting}
\label{ch3:met:headcounting}

\begin{figure}[htbp]
    \centering
    \includegraphics[width=\linewidth]{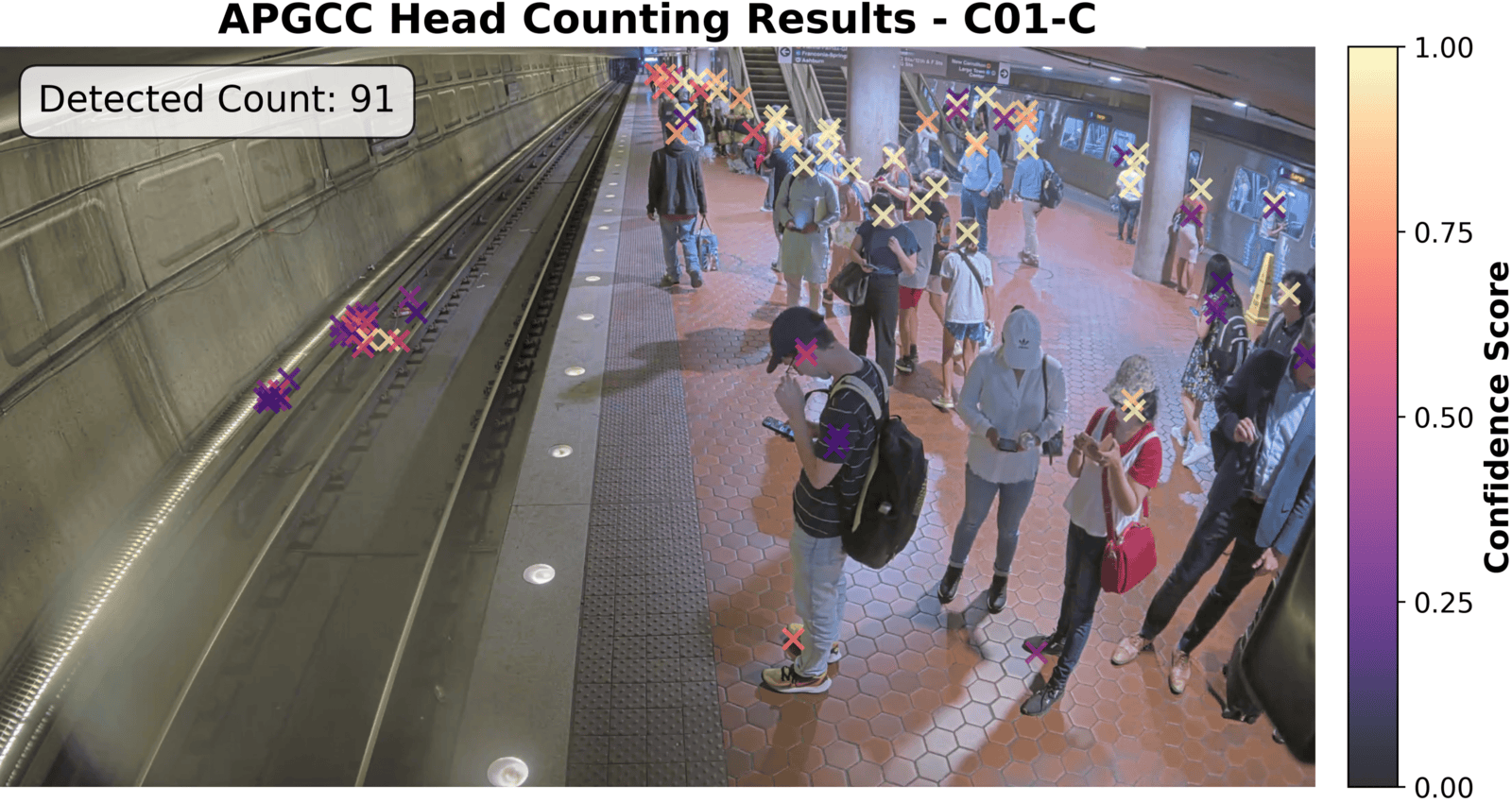}
    \caption{APGCC model head counting results with associated confidence score for C01-C \#4 \textit{before} area-of-interest masking.}
    \label{fig:apgcc_sample}
\end{figure}

In addition to object detection, one of the most salient features in overhead views of crowded scenes is the human head, which can be used to improve recall. Specifically, we employ the state-of-the-art APGCC model, pre-trained on the ShanghaiTech A datasets \cite{Zhang_2016_CVPR}. APGCC employs an auxiliary point-guidance mechanism to filter candidate head locations prior to final localization \cite{chen_improving_2024}. In \autoref{fig:apgcc_sample}, we show that the APGCC model outputs x/y-coordinates of detected heads for each image (indicated by the "x" markers). As with the object detection models, we record APGCC's maximum estimate across the three frames for each train-arrival event and, for the 15-minute bin, the mean of these maximum values.

\subsection{Crowd-Level Classification}
\label{ch3:met:scene_classification}

\subsubsection{Crowd Vision Transformer -- Crowd-ViT}
Rather than estimating exact rider counts, the \textit{magnitude} of platform occupancy often provides a sufficiently informative, potentially more robust signal for downstream prediction of concise numbers. We therefore formulate a crowd-level classification task using CCTV imagery and fine-tune a separate pre-trained Google Vision Transformer (ViT) model for each camera \cite{dosovitskiy_image_2021}.

We refer to the resulting classifier as \textit{Crowd-ViT}, denoting our camera-specific adaptation of the standard ViT image classification model, initialized with publicly available pre-trained weights. We retain the original ViT backbone, replace the original ImageNet classification head with a four-output layer corresponding to the four platform-specific crowd levels, and employ per-camera fine-tuning of the resulting model on labeled CCTV. This camera-specific training setup is motivated by the fixed viewpoints and heterogeneous fields of view across cameras. At the same time, the platform-specific label definitions account for differences in baseline occupancy levels across platforms.

For evaluation, estimates are generated on a held-out testing dataset, which we describe in the following section. We consider the \textit{median} estimated crowd level across the three sampled frames for each train-arrival event and, for the 15-minute bin aggregation, the \textit{median} estimated crowd level across all images captured in the respective bin.

\subsubsection{Crowd-Level Label Generation}
\label{ch3:met:data_prep:labels}

\begin{figure}[htbp]
    \centering
    \includegraphics[width=\linewidth]{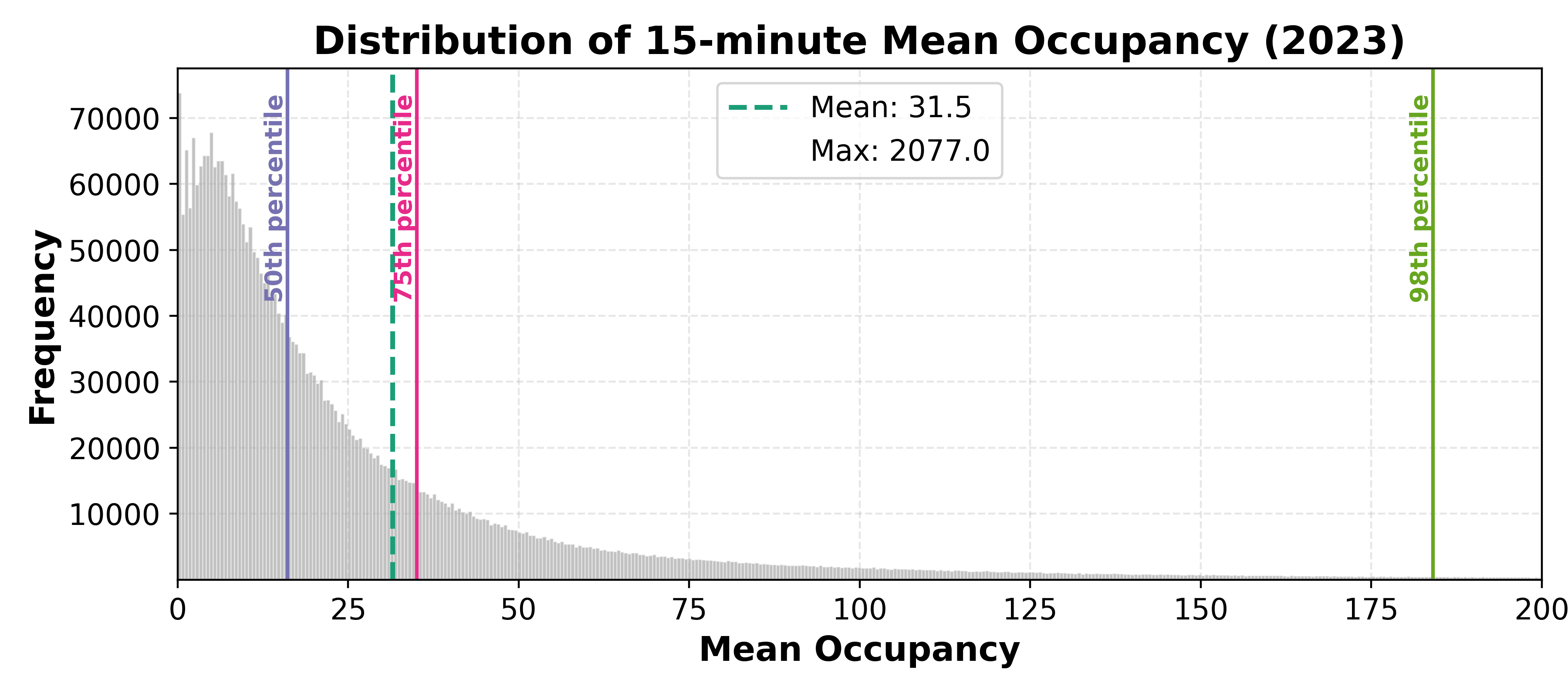}
    \caption{15-minute mean occupancy values for all platforms in the WMATA system across 2023 with 50\thh$\;$(median), 75\thh, and 98\thh$\;$percentiles evidenced.}
    \label{fig:mean_occupancy_distribution_2023}
\end{figure}

As detailed above, we use ODX's continuous per-train-arrival event occupancy estimates as target labels. For classification, however, discrete labels are required. Without normative thresholds for crowd levels, we define \textit{crowd-level} boundaries based on \textit{per-platform} occupancy distributions observed during 2023 weekday service days, excluding zero-valued occupancies. Specifically, we use the 50\thh, 75\thh, and 98\thh$\;$percentiles to delineate four crowd levels for each platform. This scheme empirically captures typical peak-hour volumes in the \textit{medium} crowd level and reserves the \textit{high} crowd level for exceptionally high occupancy values. Since percentiles are computed on non-zero occupancies, the \textit{very light} crowd level corresponds to the lowest observed non-zero volumes. 

Based on these thresholds, we define the following four crowd levels, which translate to the platform-specific values reported in \autoref{tab:crowding_occupancy_percentiles}.

\begin{table}[!ht]
    \centering
    \caption{Crowd levels and corresponding percentile ranges.}
    \label{tab:crowding_classes}
    \begin{tabular}{ll}
        \toprule
        \textbf{Crowd Level} & \textbf{Percentile Range} \\
        \midrule
        Very Light & $C_{\text{E}} \in [0^\text{th}, 50^\text{th}]$ \\
        Light      & $C_{\text{L}} \in (50^\text{th}, 75^\text{th}]$ \\
        Medium     & $C_{\text{M}} \in (75^\text{th}, 98^\text{th}]$ \\
        High       & $C_{\text{H}} \in (98^\text{th}, 100^\text{th}]$ \\
        \bottomrule
    \end{tabular}
\end{table}

\begin{table}[!ht]
    \centering
    \begin{minipage}[c]{\linewidth}
        \centering
        \caption{Crowding occupancy percentile ranges by platform.}
        \label{tab:crowding_occupancy_percentiles}
        \resizebox{\linewidth}{!}{%
    \begin{tabular}{lcccc}
    \toprule
    \textbf{Platform}
    & \textbf{0\thh--50\thh}
    & \textbf{50\thh--75\thh}
    & \textbf{75\thh--98\thh}
    & \textbf{98\thh--100\thh} \\
    \midrule

        A01-1
        & [0, 139.3]
        & (139.3, 231.8]
        & (231.8, 517.9]
        & (517.9, $\infty$] \\

        A01-2
        & [0, 138.9]
        & (138.9, 220.2]
        & (220.2, 466.8]
        & (466.8, $\infty$] \\

        C01-C
        & [0, 90.0]
        & (90.0, 145.8]
        & (145.8, 326.7]
        & (326.7, $\infty$] \\


        B01-2
        & [0, 148.8]
        & (148.8, 223.5]
        & (223.5, 434.9]
        & (434.9, $\infty$] \\


        A03-1
        & [0, 55.6]
        & (55.6, 85.0]
        & (85.0, 188.4]
        & (188.4, $\infty$] \\

        A03-2
        & [0, 51.4]
        & (51.4, 79.4]
        & (79.4, 179.9]
        & (179.9, $\infty$] \\

        F05-C
        & [0, 40.1]
        & (40.1, 69.3]
        & (69.3, 214.9]
        & (214.9, $\infty$] \\

        \hline
        \end{tabular}%
        }
    \end{minipage}
\end{table}

\subsubsection{Crowd-Level Classification Data Augmentation}
\label{ch3:met:data_prep:data_prep}

Across all platforms, distributions are heavily skewed toward lower occupancy values, even after excluding zeros. A long right-hand tail is evident, with large gaps between the median and the 98\thh$\;$percentile, highlighting the rarity and extremity of high crowding events. These patterns are consistent across the WMATA system, as can be seen from the system-wide 2023 occupancy values in \autoref{fig:mean_occupancy_distribution_2023}.

This distributional skew is also reflected in the generated dataset, as illustrated in \autoref{tab:data_per_platform}, displaying substantial imbalance of crowd levels. This imbalance can bias model training by causing overfitting to overrepresented levels and underrepresentation of rare cases, such as \textit{high crowding}. Thus, we apply stratified sampling and data augmentation techniques to address this shortcoming partially. Stratified sampling ensures that each crowd level is proportionally represented in training and testing splits, preserving crowd-level distribution. Data augmentation involves generating additional samples via image transformations to increase dataset diversity and reduce the underrepresentation of rare crowd levels.

\begin{figure}[htbp]
    \centering

    \subfigure[Per-camera training dataset distribution.
    \label{fig:classification_train_data}]{
        \includegraphics[width=\linewidth]{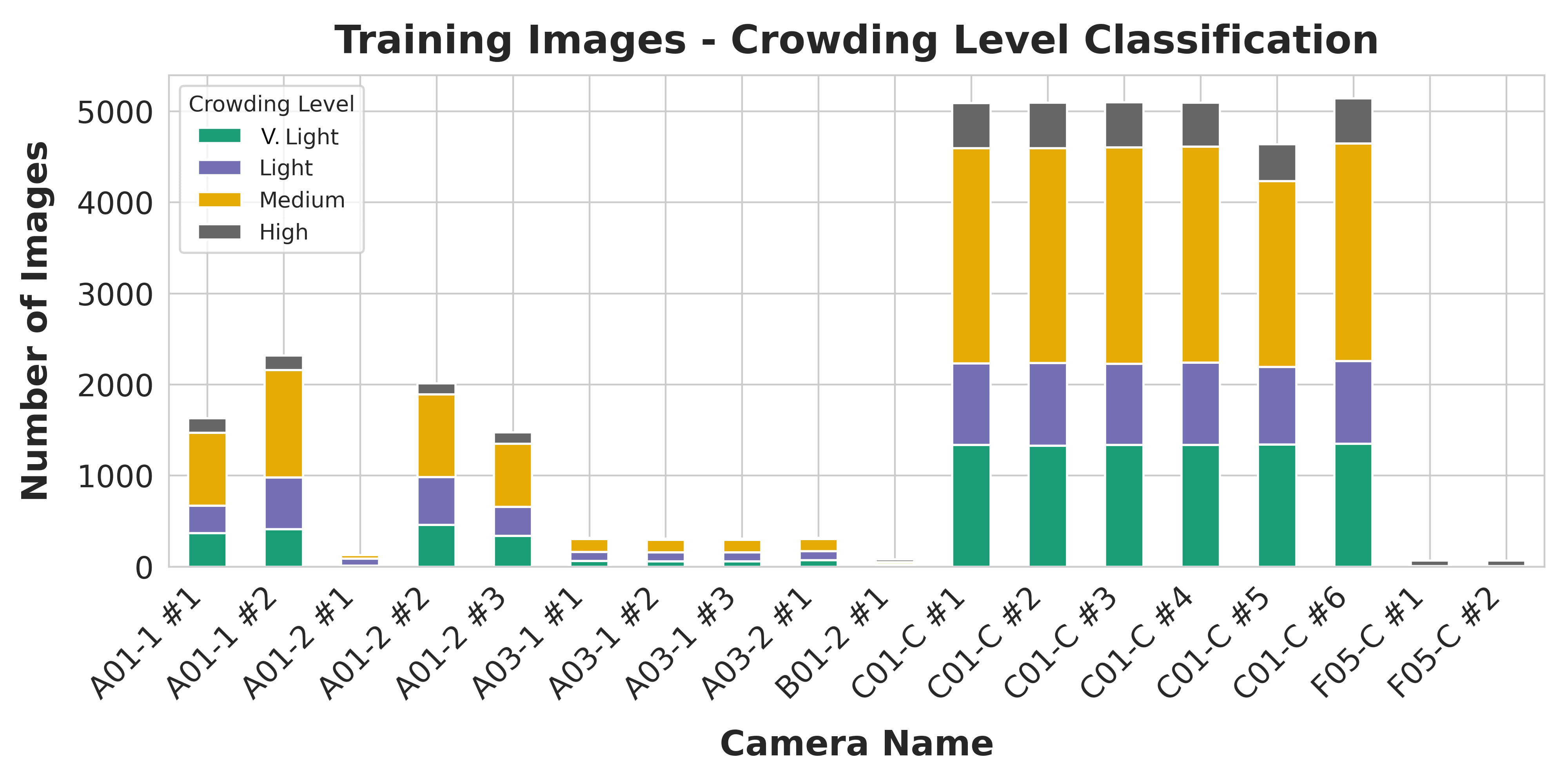}
    }
    \hfill
    \subfigure[Per-camera testing dataset distribution.
    \label{fig:classification_test_data}]{
        \includegraphics[width=\linewidth]{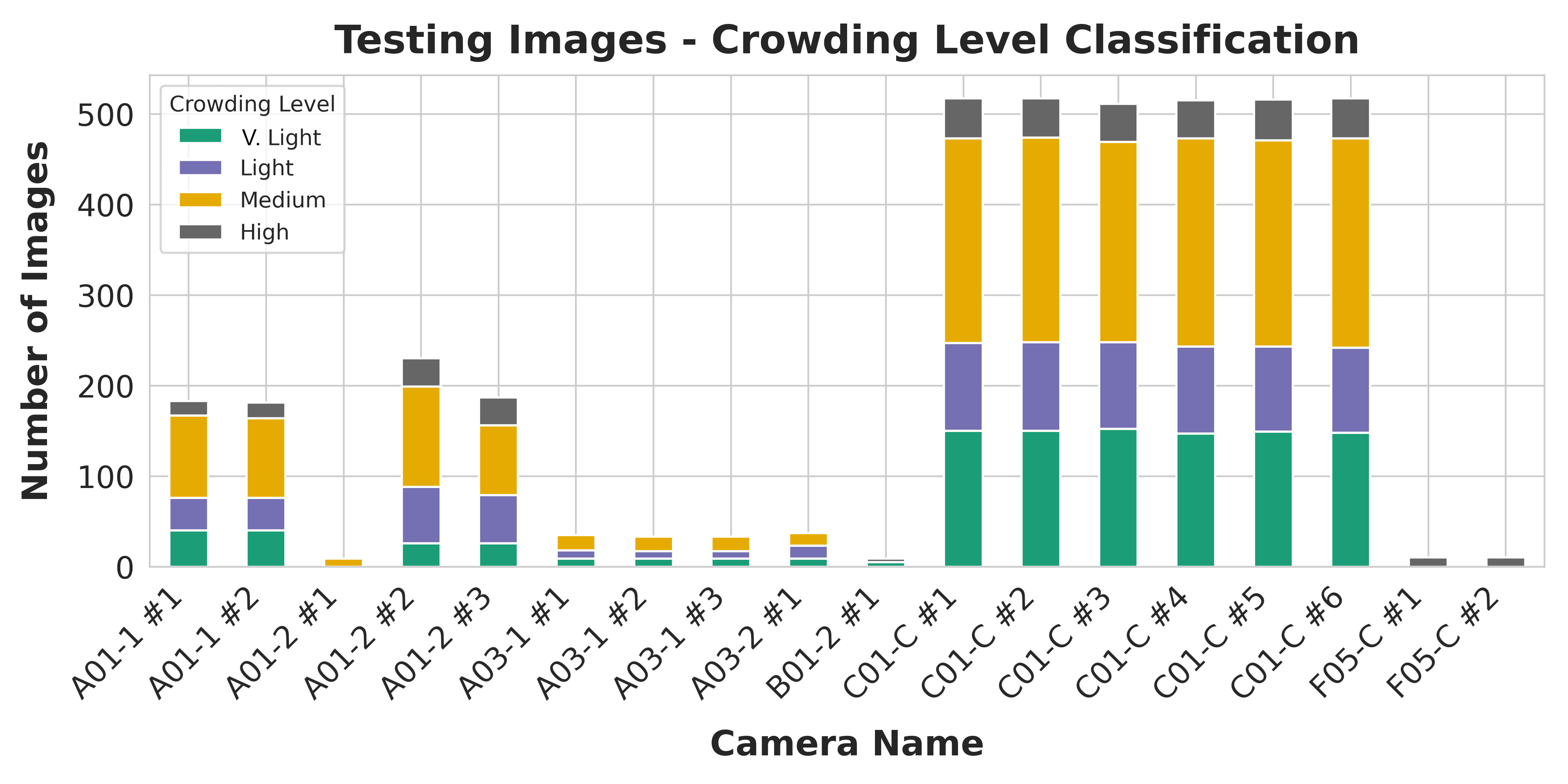}
    }

    \caption{Distributions of sampled images across crowd levels and cameras. We apply a 90\%/10\% training-testing split with stratified sampling based on crowd-level labels to ensure a similar distribution between the two sets. Images of individual train-arrival events were kept in the same set when possible.}
    \label{fig:train_test_classification_data}
\end{figure}

To train our Crowd-ViT models, we apply a 90\%/10\% training/testing split stratified by crowd level. Where possible, we restrict the testing set to train-arrival events that are jointly captured by all cameras on a given platform, thereby avoiding leakage of visually correlated images. However, no set of images from a single train-arrival event present in the training set ever appears in the test set. This results in 1,599 jointly captured train-arrival events (32,498 images) for training and 179 (3,624 images) for testing. An additional 6,684 training and 426 testing images come from train-arrival events captured by only one camera. For A01-2, where no jointly captured train-arrival events were available, train-arrival events were randomly split while maintaining crowd-level stratification and respecting the training/testing split. \autoref{fig:train_test_classification_data} shows the resulting distributions across crowd levels and cameras.

\begin{figure}[htbp]
    \centering
    \includegraphics[width=\linewidth]{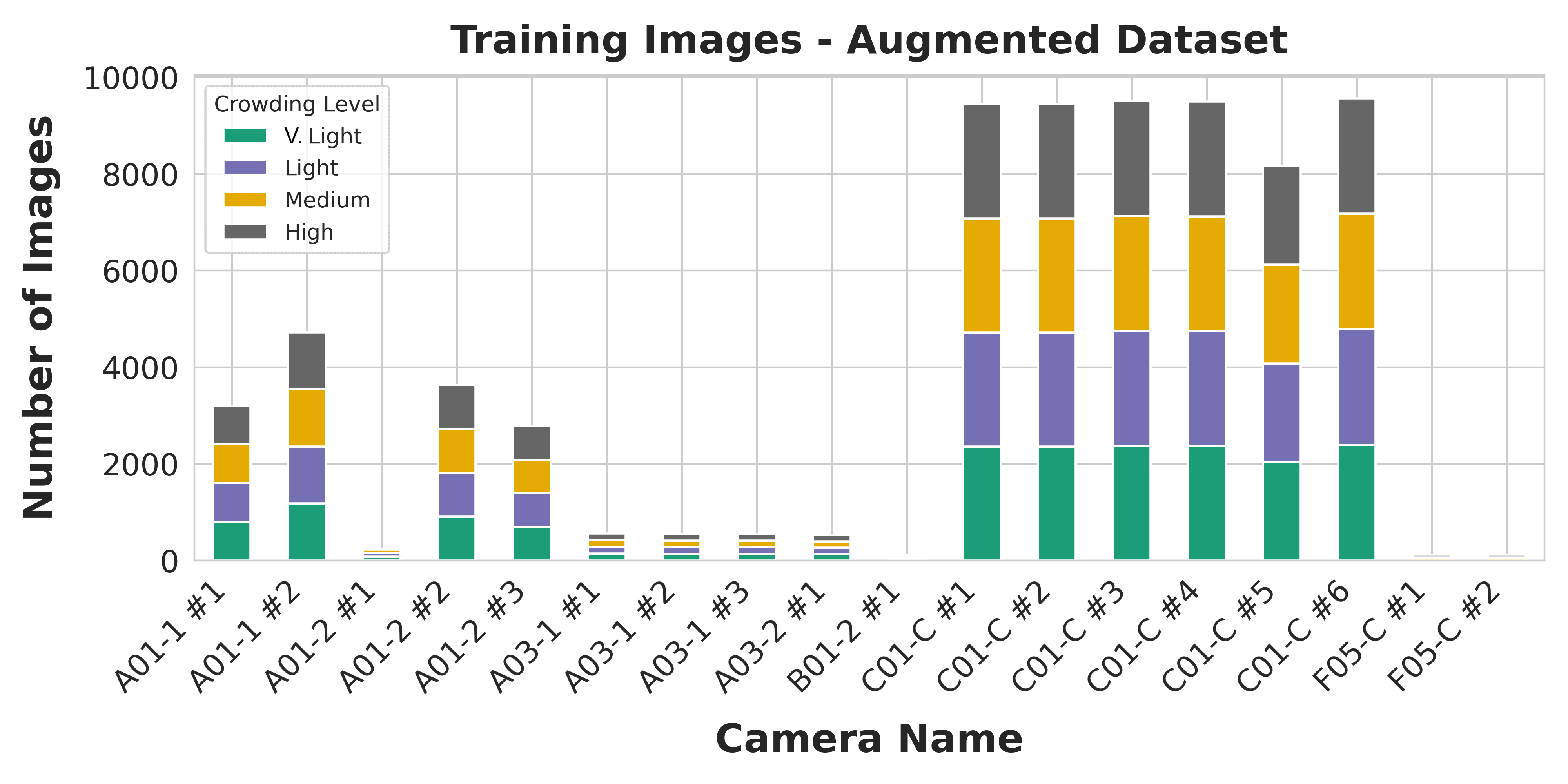}
    \caption{Crowd-ViT classification training dataset balanced through augmentation.}
    \label{fig:classification_augmented_train_data}
\end{figure}

Furthermore, we apply data augmentation to the training dataset. For the \textit{very light} level, where variation between images is minimal, na\"ive image repetition suffices. For the remaining levels, we apply targeted data augmentation using random brightness and contrast shifts, color jitter, rotations up to 15°, Gaussian noise, and motion blur, each applied with 30\% probability. This results in a balanced training set across all crowd levels, as illustrated in \autoref{fig:classification_augmented_train_data}.

\begin{figure}[htbp]
    \centering
    \includegraphics[width=0.3\textwidth]{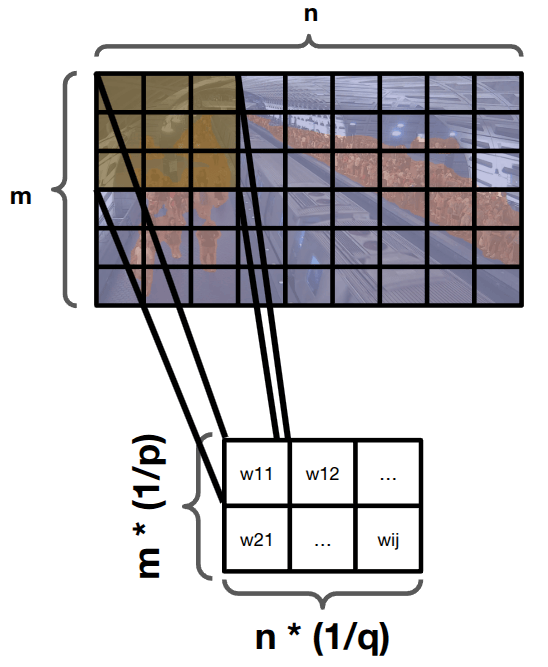}
    \caption{Visualization of segmentation map, mean pooling, and multiplication with weight matrix. Underlying is the DeepLabV3-generated binary segmentation map (person-class/background pixels) of size $m \times n$. The learned weight matrix is then applied to each pooled region.}
    \label{fig:weight-process}
\end{figure}

\subsection{Crowd Segmentation}
\label{ch3:met:crowdsegmentation}


Finally, we apply semantic image segmentation to estimate the area of the image that riders occupy in a given CCTV frame, serving as a proxy for occupancy estimation. We employ the DeepLabV3 model (model \textit{ResNet101}, 58.6 million parameters) \cite{vainf_deeplabv3, chen_deeplab_2017} pre-trained on the Pattern Analysis, Statistical Modeling and Computational Learning Visual Object Classes (PASCAL VOC) 2012 dataset \cite{everingham_pascal_2015}. The model produces binary segmentation maps identical in size to the input images, assigning value \textit{one} to pixels classified as \textit{class 15} (\textit{person}) in the PASCAL VOC taxonomy and \textit{zero} otherwise.

From these maps, we extract three representations. First is the raw pixel count of detected person-class regions. Second, the ratio of human-class pixels to total image area. Third, an estimated platform occupancy count is obtained by multiplying the segmentation maps by a learned, camera-specific calibration weight map. For the latter, we formulate a Ridge Regression (RR)-based model that produces pixel-weighting matrices, enabling efficient and interpretable mappings from segmentation outputs to crowd estimates.

\subsubsection{Segmentation Map Weighting}
\label{ch3:met:ridge_regression}

Our dataset benefits from fixed camera viewpoints and consistent platform view angles. The underlying passenger behavior, entering through designated platform access points and queuing near train doors, generates a stable visual structure. Since all images are sampled during train-arrival events, the spatial distribution of queuing serves as a reliable proxy for occupancy levels.

Object detection and classification models yield directly interpretable outputs such as counts or labels, but segmentation methods produce dense, pixel-level maps requiring post-hoc calibration. To bridge the gap between these pixel activations and the ODX-estimated platform occupancies, we introduce the following RR formulation:
\begin{align*}
    \min_{\{\mathbf{w}_c\}_{c\in C}}
    &\quad
    \sum_{c\in C}\sum_{a\in A^c}\sum_{i\in I^a_c}
    \Biggl(
        y^i_a
        -
        \sum_{j=1}^{\frac{m}{p}}
        \sum_{k=1}^{\frac{n}{q}}
        \mathbf{P}^i_{a,c}[j,k]\mathbf{w}_c[j,k]
    \Biggr)^2 \\
    &\quad
    + \lambda
    \sum_{c\in C}
    \sum_{j=1}^{\frac{m}{p}}
    \sum_{k=1}^{\frac{n}{q}}
    \mathbf{w}_c[j,k]^2, \\[1ex]
    \text{s.t.:} 
    &\quad
    \mathbf{P}^i_{a,c}[j,k]
    = 
    \frac{1}{p\,q}
    \sum_{u=0}^{p-1}
    \sum_{v=0}^{q-1}
      \mathbf{X}^c_i[jp+u,\,kq+v]
    \\[0.5ex]
    &\quad
    \forall\,c\in C,\;a\in A^c,\;i\in I^a_c,\;j,k, \\[0.5ex]
    &\quad \mathbf{w}_c[j,k]\geq 0
    \quad \forall\,c\in C,\; j,k,\\[0.5ex]
    &\quad
    \mathbf{P}^i_{a,c}\in\mathbb{R}^{\frac{m}{p}\times\frac{n}{q}},
    \quad
    \mathbf{w}_c\in\mathbb{R}^{\frac{m}{p}\times\frac{n}{q}}, \\[0.5ex]
    &\quad
    p,q\text{ divide }m,n,\quad
    \lambda>0,\quad
    y\in\mathbb{R}^{|I|}_{\geq 0},\quad
    \\[0.5ex]
    &\quad X^c_i\in\{0,1\}^{m\times n}, \\[0.5ex]
    &\quad
    c\in C\text{ (cameras)},\\[0.5ex]
    &\quad
    a\in A^c\text{ (train-arrival events captured by }c\! ),\\[0.5ex]
    &\quad i\in I^a_c\text{ (images of event }a\text{ captured by }c\!).
\end{align*}
By construction, this is a convex ridge regression over mean-pooled segmentation features. Specifically, for each camera $c\in C$, we index train-arrival events by $a\in A^c$ and the corresponding images by $i\in I^a_c$. The term $y^i_a$ denotes the ground‐truth ODX‐based platform occupancy estimate for the $i$-th image of event $a$, as captured by camera $c$. The binary segmentation map $\mathbf{X}^c_i\in\{0,1\}^{m\times n}$ is produced by the DeepLabV3 model to indicate regions of class \textit{person} in the original image. In contrast, $\mathbf{P}^i_{a,c}\in\mathbb{R}^{\frac{m}{p}\times\frac{n}{q}}$ is obtained by applying a \textbf{mean}‐pooling operation over non‐overlapping $p\times q$ blocks of $\mathbf{X}^c_i$, illustrated in  \autoref{fig:weight-process}, thereby reducing its size. The indices \(j\in\{1,\ldots,m/p\}\) and \(k\in\{1,\ldots,n/q\}\) denote the row and column, respectively, of the pooled spatial grid. The camera‐specific weight matrix $\mathbf{w}_c\in\mathbb{R}^{\frac{m}{p}\times\frac{n}{q}}$ shares the same shape as this pooled segmentation map $\mathbf{P}^i_{a,c}$. 

Here, the pooled regions are fixed cells within an image, rather than a physical segment of the platform. The pooling parameters \(p\) and \(q\) determine their granularity, with each cell containing the mean person-class activation within one non-overlapping \(p\times q\) pixel block, yielding an \((m/p)\times(n/q)\) spatial grid. 

\begin{figure}[!ht]
    \centering

    \subfigure[Learned weighting map.
    \label{fig:calibration_results:weights}]{
        \includegraphics[width=0.3\linewidth]{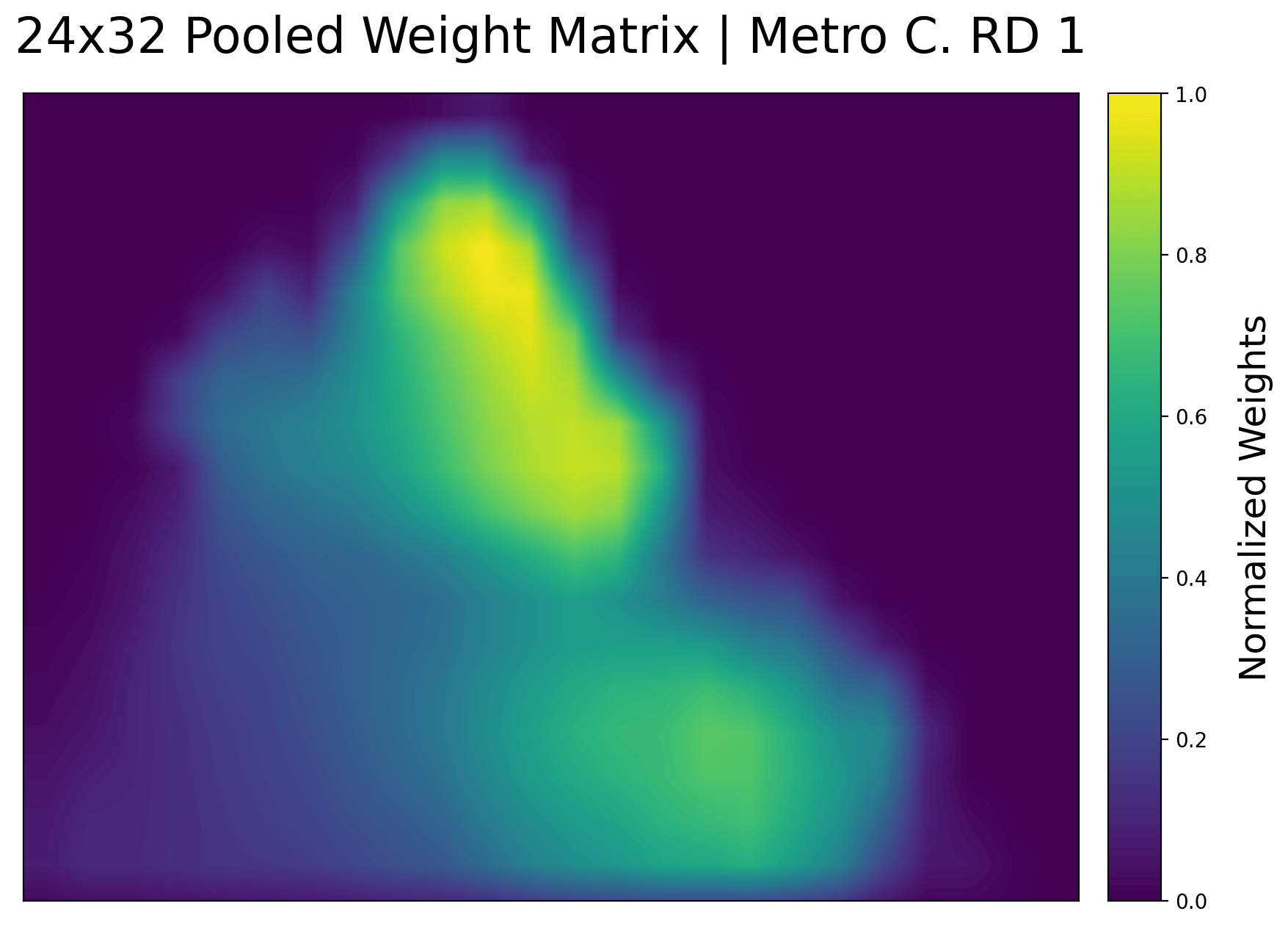}
    }
    \hfill
    \subfigure[Base image.
    \label{fig:calibration_results:base_image}]{
        \includegraphics[width=0.27\linewidth]{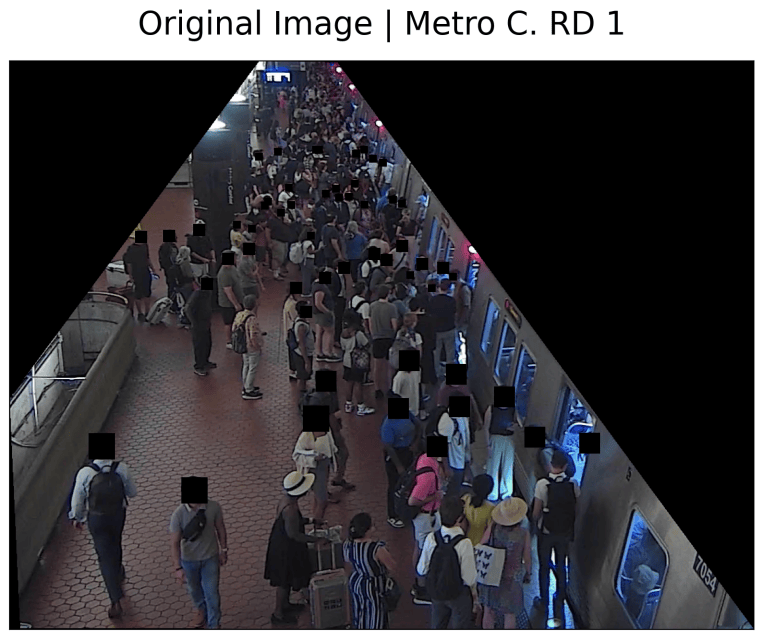}
    }
    \hfill
    \subfigure[Overlay.
    \label{fig:calibration_results:overlay}]{
        \includegraphics[width=0.34\linewidth]{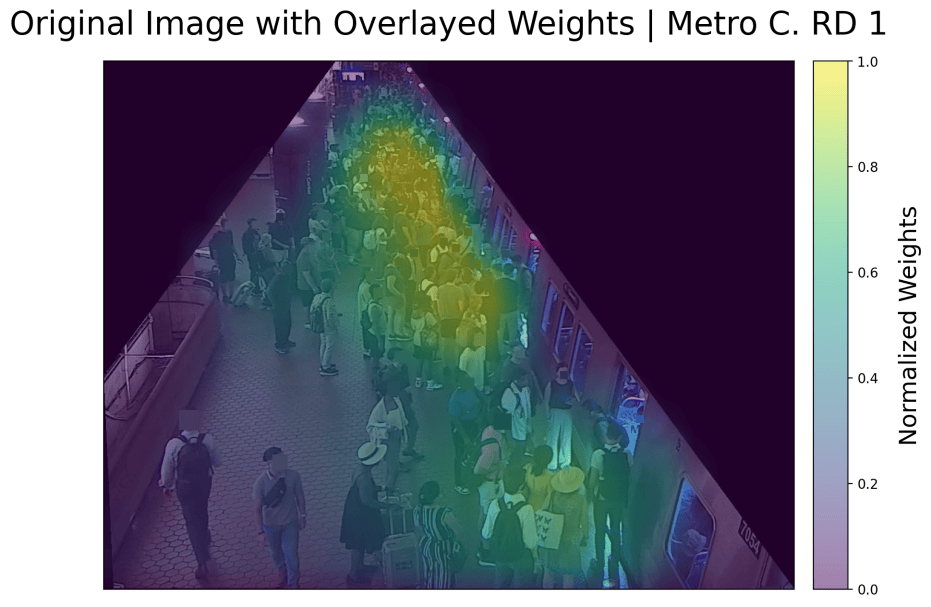}
    }

    \caption{Sample of how a learned segmentation map calibration relates to the main crowding areas in a CCTV image. This calibration map is used to compute platform occupancy estimates from segmentation maps provided by a DeepLabV3 semantic segmentation model.}
    \label{fig:calibration_results}
\end{figure}

The objective in the formula above is then to minimize the squared residuals between the ground‐truth occupancy $y^i_a$ and the weighted sum:
\begin{equation}
    \sum_{j=1}^{\frac{m}{p}}\sum_{k=1}^{\frac{n}{q}} \mathbf{P}^i_{a,c}[j,k]\,\mathbf{w}_c[j,k],
\end{equation}
with an additional L2 regularization term $\lambda\sum_{c\in C}\sum_{j,k}\mathbf{w}_c[j,k]^2$ to discourage large weights. We solve this convex program using the ECOS solver \cite{bib:Domahidi2013ecos} via the Python CVXPY interface \cite{cvxpy}.

\begin{figure}[htbp]
    \centering
    \includegraphics[width=\linewidth]{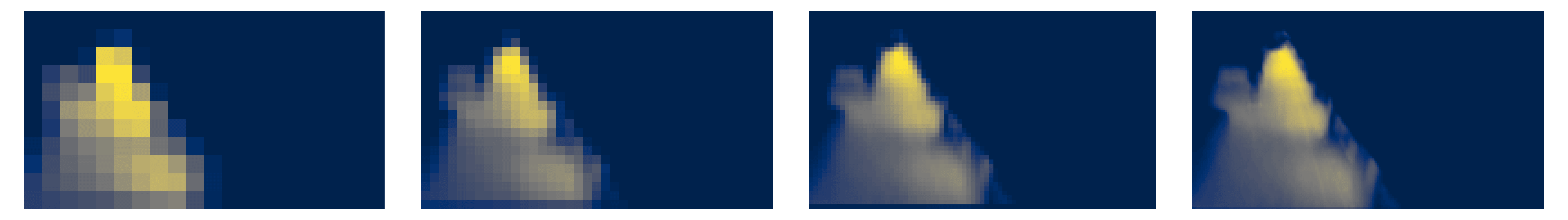}
    \caption{Learned weight maps with increasing resolution as pixel-pooling size is reduced (from left to right). Smaller pooling sizes reveal more platform features, with higher weights near train entrances and farther from the camera.}
    \label{fig:weight-maps}
\end{figure}

Once the optimization yields each camera's weight matrix $\mathbf{w}_c$, we compute a real-valued estimate of the platform occupancy by element‐wise multiplying the learned weight matrix with the pooled DeepLabV3 segmentation maps $\mathbf{P}^i_{a,c}$ in the testing dataset. Visually, $\mathbf{w}_c$ assigns higher weights to regions corresponding to areas closer to the train and farther from camera $c$, as demonstrated in \autoref{fig:calibration_results}. Furthermore, \autoref{fig:weight-maps} illustrates how different choices of pooling granularity $(p,q)$ affect the resolution of $\mathbf{w}_c$, preserving the overall spatial weighting structure while varying detail.

\subsection{Gradient Boosting Decision Trees}
\label{ch2:rel:lgbm_intro}

CV-derived counts often suffer from scale differences. Thus, to map CV-derived features to platform occupancy estimates, we use Light Gradient Boosted Machines (LGBM) \cite{NIPS2017_6449f44a}, a non-linear ensemble method that sequentially builds decision trees, each correcting residuals from the previous iteration. LGBM's histogram-based training, leaf-wise growth, and native support for regularization, missing values, and parallelization make it well-suited for high-dimensional input data \cite{NIPS2017_8a20a862}.

We train a separate model for each CV pipeline to predict platform occupancy at the per-train-arrival level. For each arrival, the three sampled frames ($-5$s, $0$s, $+5$s) from each available camera are compressed into a single scalar feature per camera. Specifically, we retain the maximum person or head count for detection and head-counting models (YOLOv11, RT-DETRv2, APGCC), the maximum calibrated segmentation estimate for DeepLabV3, and the median ordinal crowd-level prediction for Crowd-ViT. Concatenating these per-camera features yields one feature vector per arrival, from which the model produces a continuous occupancy estimate. Its ability to handle missing inputs allows the inclusion of arrivals without full camera coverage. We then average the per-arrival estimates within each 15-minute interval to assess aggregate-level performance.

\subsection{Quantifying Measurement Error and Latent Process Noise}

ODX-inferred platform occupancies used as CCTV model labels contain uncertainty from the underlying passenger-assignment and queuing procedures. To characterize this uncertainty, we model the contiguous time series of 15-minute mean ODX occupancies within a linear Gaussian SSM, treating \textit{true} platform occupancy as an unobserved state and decomposing unexplained variation into \emph{latent process noise} and \emph{measurement noise}. The latent process follows auto-regressive ARX(1)-type dynamics, with bin-aggregated AFC inflows and outflows and historical occupancies as exogenous inputs. We fit the model by Maximum Likelihood Estimation (MLE) using Kalman filter recursions on 2023 ODX data for each studied platform \cite{durbin_time_2012}. The resulting in-sample one-step-ahead estimates are used for model-based variance decomposition rather than operational forecasting.

\subsubsection{SSM Specification}

Let $X_t$ denote the latent, i.e. \textit{unobserved}, \textit{true} mean platform occupancy during the 15-minute interval $(t-1,t]$, and let $Y_t$ denote the corresponding ODX-reported value. Let $\boldsymbol{Z}_t$ contain the exogenous predictors, AFC-based entries and exits in $(t-1,t]$ and historic occupancy features (same-time previous day and previous 1,2,3 weeks). The latent dynamics and observation process are then specified as:
\begin{align}
    X_t &= \alpha + \phi X_{t-1} + \beta^\top \boldsymbol{Z}_t + u_t, 
     &u_t \overset{i.i.d.}\sim \mathcal{N}(0,\sigma_u^2),\\[1mm]
    Y_t &= X_t + e_t, 
     &e_t \overset{i.i.d.}\sim \mathcal{N}(0,\sigma_e^2),
\end{align}
where $u_t$ represents unexplained temporal variation in the latent occupancy process and $e_t$ represents model-implied ODX observation noise under the assumed SSM. The parameter vector $(\alpha,\phi,\beta,\sigma_u^2,\sigma_e^2)$ is computed by the aforementioned MLE. Using these estimates, we apply the Kalman smoother to obtain the smoothed latent state $\widehat{X}_{t\mid T} = \mathbb{E}[X_t \mid Y_1,\dots, Y_T]$, which provides a reconstructed latent time series of mean occupancies. The inferred standard deviations $\sigma_u$ and $\sigma_e$ characterize the model-implied partition of unexplained variability into latent process variability and ODX observation noise, respectively. Both noise terms are assumed to be zero-mean Gaussian and independent over time, i.e., independently and identically distributed (i.i.d.).

\subsubsection{Validation of the Fit}

The resulting variance decomposition is inherently \textit{model-based}, i.e., it reflects how the ARX(1) SSM attributes unexplained variation to latent process noise versus measurement noise. The standardized one-step-ahead innovations are approximately Gaussian but exhibit statistically significant autocorrelation according to the Ljung-Box test, indicating that the simple ARX(1) SSM does not capture all temporal structure in the series. Accordingly, the estimated $\sigma_e$ values should be interpreted as order-of-magnitude measures of ODX observation noise under this specification rather than exact calibration errors.

\subsection{Performance Evaluation}
\label{ch3:met:performance_evaluation}

With further insight into the structural errors in our ODX-based labels, we proceed to quantify the effectiveness of our various CV approaches for estimating platform occupancy across the different granularities we consider, namely, per-image, per-train-arrival event, and per-15-minute bin. To formalize this task, we assume a model $f$ which maps a feature vector $\boldsymbol{x}$, derived from CCTV imagery, to an estimated occupancy $\hat{y}$, i.e. $\hat{y} = f(\boldsymbol{x})$. To evaluate the performance of $f$, we use the following metrics. For the occupancy estimation task, we distinguish:

\begin{itemize}
    \item Per-train-arrival event estimates $\hat{y}_a$ for each train-arrival event $a \in A$.
    \item Per-image estimates $\hat{y}^i_a$ for each image $i \in I_a$ associated to train $a \in A^c$ captured by camera $c \in C$.
    \item Per-interval estimates $\hat{\bar{y}}_t$ for each 15-minute bin $t \in T$.
\end{itemize}

Since ODX provides per train-arrival event occupancy estimates $y_a$, these serve as targets for $\hat{y}_a$ and $\hat{y}^i_a$. For aggregated estimates $\hat{\bar{y}}_t$, the target is the 15-minute mean occupancy:
\begin{equation}
    \begin{aligned}
        \bar{y}_t
        &= \frac{1}{|A^c_t|} \sum_{a \in A^c_t} y_a, \\
        A^c_t
        &= \{a \in A^c \mid
        \text{train } a \text{ arrived during time bin } t\}.
    \end{aligned}
\end{equation}
First, we report a \textit{standardized agreement score}, denoted $\tilde{R}^2$, to characterize similarity in variation independently of differences in absolute scale and offset between vision-based estimates and ODX-derived reference occupancies. We independently standardize $y$ and $\hat{y}$ and compute
\begin{equation}
    \tilde{R}^2 = R^2\left( \text{z}(y), \text{z}(\hat{y}) \right).,
\end{equation}
where the z-transformation is computed using
\begin{equation}
    \text{z}(x) = \frac{x - \mu_x}{\sigma_x},
\end{equation}
with $\mu_x$ and $\sigma_x$ denoting the mean and standard deviation of the variable being standardized.

Under this definition, $\tilde{R}^2=2r-1$, where $r$ is the Pearson correlation coefficient. Thus, $\tilde{R}^2=1$ denotes perfect positive linear agreement after standardization, while negative values indicate $r<0.5$.
We therefore use $\tilde{R}$ to characterize standardized agreement, while the following error metrics quantify absolute occupancy-estimation error.

Second, to evaluate the error between the ODX-derived reference occupancy $y$ and our estimates $\hat{y}$, we employ the 95\thh$\;$percentile Absolute Error (95\thh$\;$AE). Here, the 95\thh$\;$AE corresponds to the 95\thh$\;$percentile of the absolute error distribution, indicating the error magnitude that is not exceeded by 95\% of observations. The formulation, where $\eta$ denotes the percentile, can be written as:
\begin{equation}
    \text{AE 95}^{\text{th}} = \eta_{.95} \left( \left| y - \hat{y} \right| \right).
\end{equation}
Third, we define a custom weighted Mean Absolute Error (wMAE), which more severely penalizes estimation errors during periods of atypical platform occupancy by incorporating temporal context via a z-score-based weight.

Let $t \in \{1, \ldots, T\}$ index each 15-minute bin in the testing set, and let $y_t$ and $\hat{y}_t$ denote the observed and estimated occupancy values at time $t$, respectively. Each time-step $t$ is assigned to one of $B = 4 \times 24 \times 7 = 672$ 15-minute bins in the weekly cycle, indexed by $b(t) \in \{1, \ldots, B\}$. Let $\mu_b$ and $\sigma_b$ denote the historical mean and standard deviation of occupancy for bin $b$, computed across all weeks in 2023. To avoid division by zero and excessive weights for bins with very light historical variance, we introduce a small positive constant $\epsilon > 0$ as a lower bound on the standard deviation. For each $t$, we define a z-score-based weight as follows, where the term inside the norm represents the traditional z-score computation:
\begin{equation}
    w_t = 1 + \left| \frac{y_t - \mu_{b(t)}}{\max(\sigma_{b(t)}, \epsilon)} \right|.
\end{equation}
The weighted Mean Absolute Error (wMAE) is then defined as:
\begin{equation}
    \text{wMAE} = \frac{1}{\sum_{t=1}^T w_t} \sum_{t=1}^T w_t \cdot |y_t - \hat{y}_t|.
    \label{ch3:wmae_definition}
\end{equation}
Note that the unweighted Mean Absolute Error (MAE) is implicitly defined with $w_t=1 \;\forall t\in T$. This formulation increases the penalty for estimation errors during periods of high deviation, such as unusual morning or afternoon peaks, operational disruptions, or special events. The presented formulation for $w_t$ ensures that $w_t\geq1\;\forall t\in T$, preserving numerical stability.

\section{Experiments \& Results}
\label{ch3:experiments_and_results}

This section evaluates CV models for estimating ODX-derived reference occupancy from CCTV imagery at the per-image, per-train-arrival, and per-15-minute-bin levels. We first present individual model performance, followed by an analysis of LGBM-based aggregation across cameras per train-arrival event. Lastly, we assess 15-minute bin accuracy using LGBM outputs. A special focus is laid on Metro Center platforms A01-1/2 and C01-C due to high data availability. Performance differences between side and center platforms are also examined. All experiments presented were run on an MIT SuperCloud instance with one Nvidia V100 Graphics Processing Unit (GPU), 20 Intel Xeon-G6 cores, and 170GB of RAM.

\begin{figure}[!ht]
    \centering
    \subfigure[\textit{Per-image:} Each data point corresponds to a single CCTV frame captured. This produces a higher-resolution but noisy mapping between object detection counts and ODX-derived reference occupancy.
    \label{fig:two_platforms_per_image}]{
        \includegraphics[width=\linewidth]{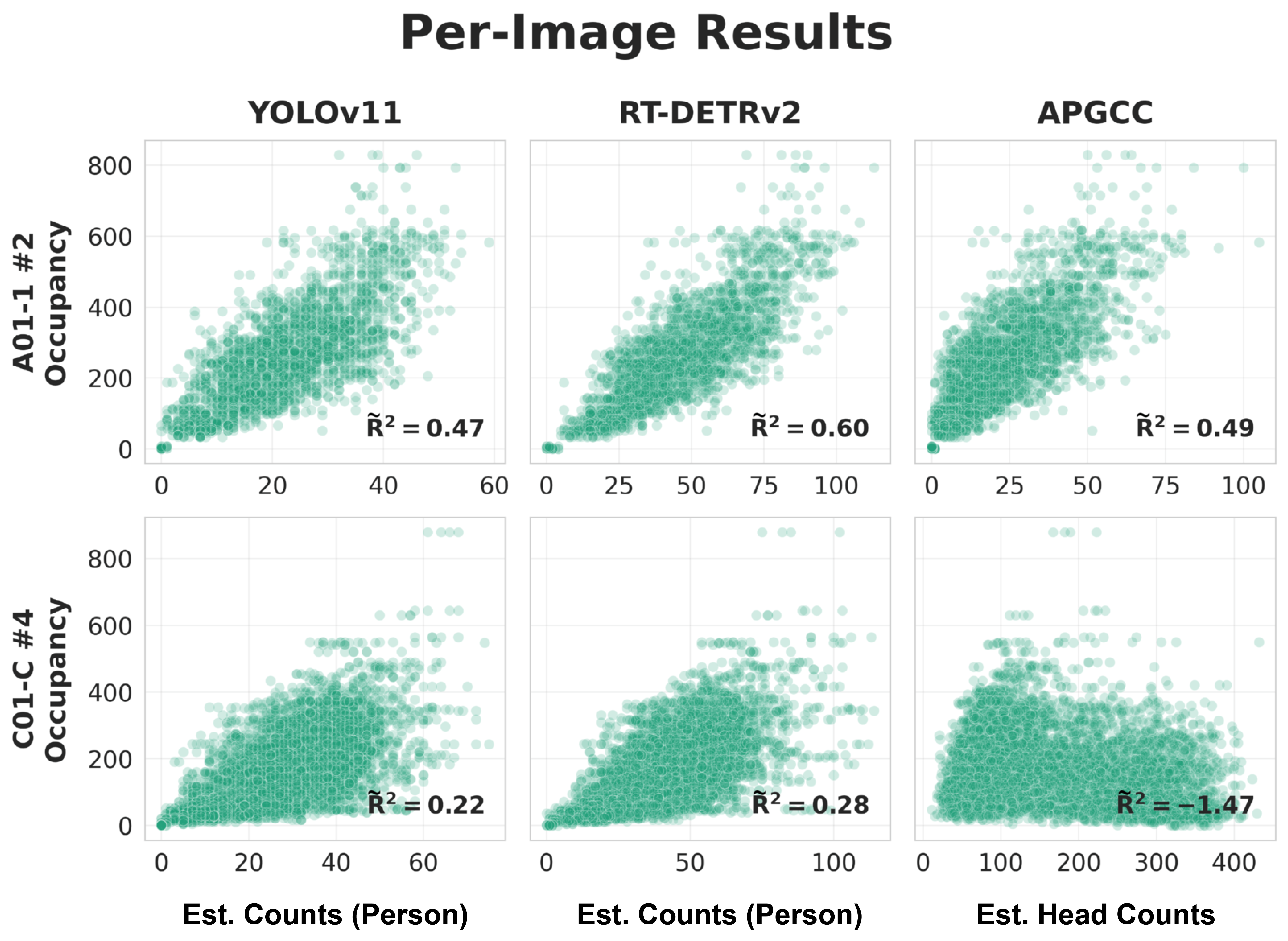}
    }
    \hfill
    \subfigure[\textit{Per-train-arrival event:} For each train-arrival event, the maximum detection count across frames is used. This approach aggregates the signal from \autoref{fig:two_platforms_per_image}, yielding a clearer relationship between observed counts and ODX-derived reference occupancy.
    \label{fig:two_platforms_per_train}]{
        \includegraphics[width=\linewidth]{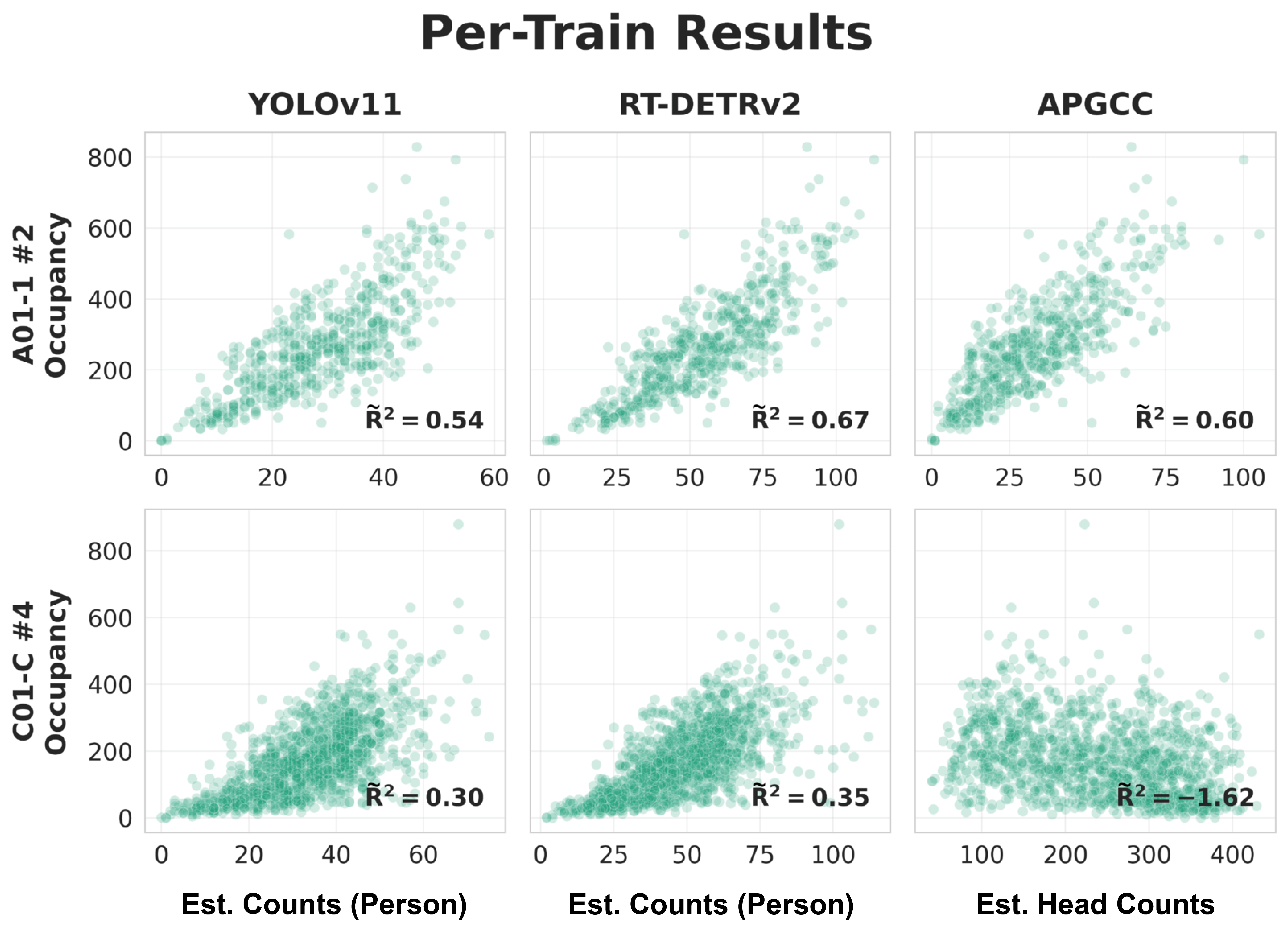}
    }
    \caption{Estimated counts of class \textit{Person} (YOLOv11 / RT-DETRv2) and estimated head counts (APGCC) plotted against ODX-derived reference occupancies. \textbf{Top:} Metro Center Red line platform A01-2 (\textit{side platform}). \textbf{Bottom:} Metro Center Blue/Orange/Silver line platform C01-C (\textit{center platform}).}
    \label{fig:two_platforms_object_detection_models}
\end{figure}

\subsection{Object Detection}
\label{ch3:results:object_detection}

Applied to the full CCTV dataset, we evaluate three pre-trained, detection-based models: YOLOv11, RT-DETRv2, and the head-counting model APGCC. \autoref{fig:two_platforms_object_detection_models} shows results from one representative camera on a side and one on a center platform.

Per-image estimates in \autoref{fig:two_platforms_per_image} show positive correlation of the estimated counts of the \textit{Person} class and the estimated head counts with ODX-derived reference occupancy across all models. Side platforms yield higher $\tilde{R}^2$ scores, likely due to elevated camera angles and reduced directional passenger mixing. RT-DETRv2 with SAHI shows the highest $\tilde{R}^2$ value, capturing a broad range of occupancies. APGCC performs reasonably on side platforms but fails on center platforms due to systematic false positives, where floor tiles and tracks are often misclassified as heads, resulting in negative $\tilde{R}^2$ values (see \autoref{fig:apgcc_sample}). All models undercount in high-density scenes, with maximum observed counts of 96 (YOLOv11 + SAHI), 199 (RT-DETRv2 + SAHI), and 455 (APGCC).

Aggregating detections per train-arrival event improves standardized agreement and reduces estimation dispersion, as shown in \autoref{fig:two_platforms_per_train}. This temporal aggregation also reduces noise, narrowing the dispersion of estimates. RT-DETRv2 retains a high agreement score, while YOLOv11 continues to greatly undercount, with an underestimation following a pseudo-exponential trend at mid-range occupancies. APGCC overestimates under low occupancy on side platforms and fails on center platforms, resulting in negative $\tilde{R}^2$ values.

\subsection{Crowd-Level Classification}
\label{ch3:results:classification}

Fine-tuning per-camera Crowd-ViT models on our augmented, training dataset balanced across crowd levels yields strong training performance but limited generalization. On the training set, per-image accuracy reaches 89.15\%, with high F1 scores across all crowd levels, including \textit{very light} (0.93), \textit{medium} (0.92), and \textit{high} (0.94). Aggregating estimates to the per-train-arrival event level further improves accuracy to 92.19\%, with F1 scores exceeding 0.94 for all crowd levels except \textit{light} (0.73).

In contrast, performance on the testing dataset drops substantially, with 60.96\% per-image accuracy and low F1 scores for the under-represented \textit{light} (0.24) and \textit{high} (0.32) levels. The model performs best on the more frequent \textit{very light} (0.74) and \textit{medium} (0.70) levels. Per-train-arrival aggregation improves testing accuracy to 63.42\%, with slight gains in F1 for \textit{very light} (0.77) and \textit{medium} (0.72). Overall, accuracy is higher on side platforms (69.05\%) than on center platforms (61.76\%) across all studied platforms.

\begin{figure}[!ht]
    \centering
    \subfigure[Training set - center platforms.
    \label{fig:conf_matrix_train_center}]{
        \includegraphics[width=0.45\linewidth]{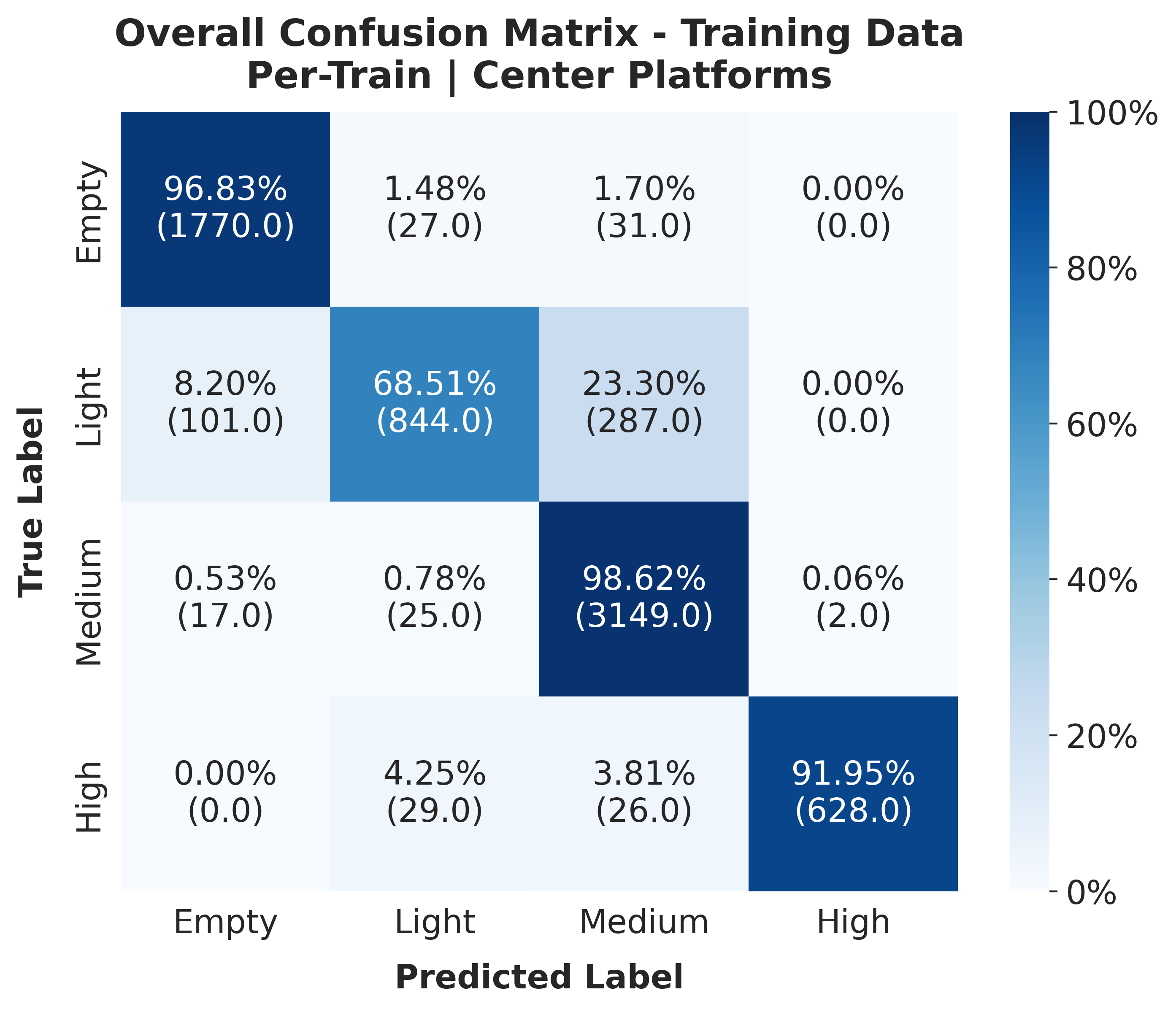}
    }
    \hfill
    \subfigure[Training set - side platforms.
    \label{fig:conf_matrix_train_side}]{
        \includegraphics[width=0.45\linewidth]{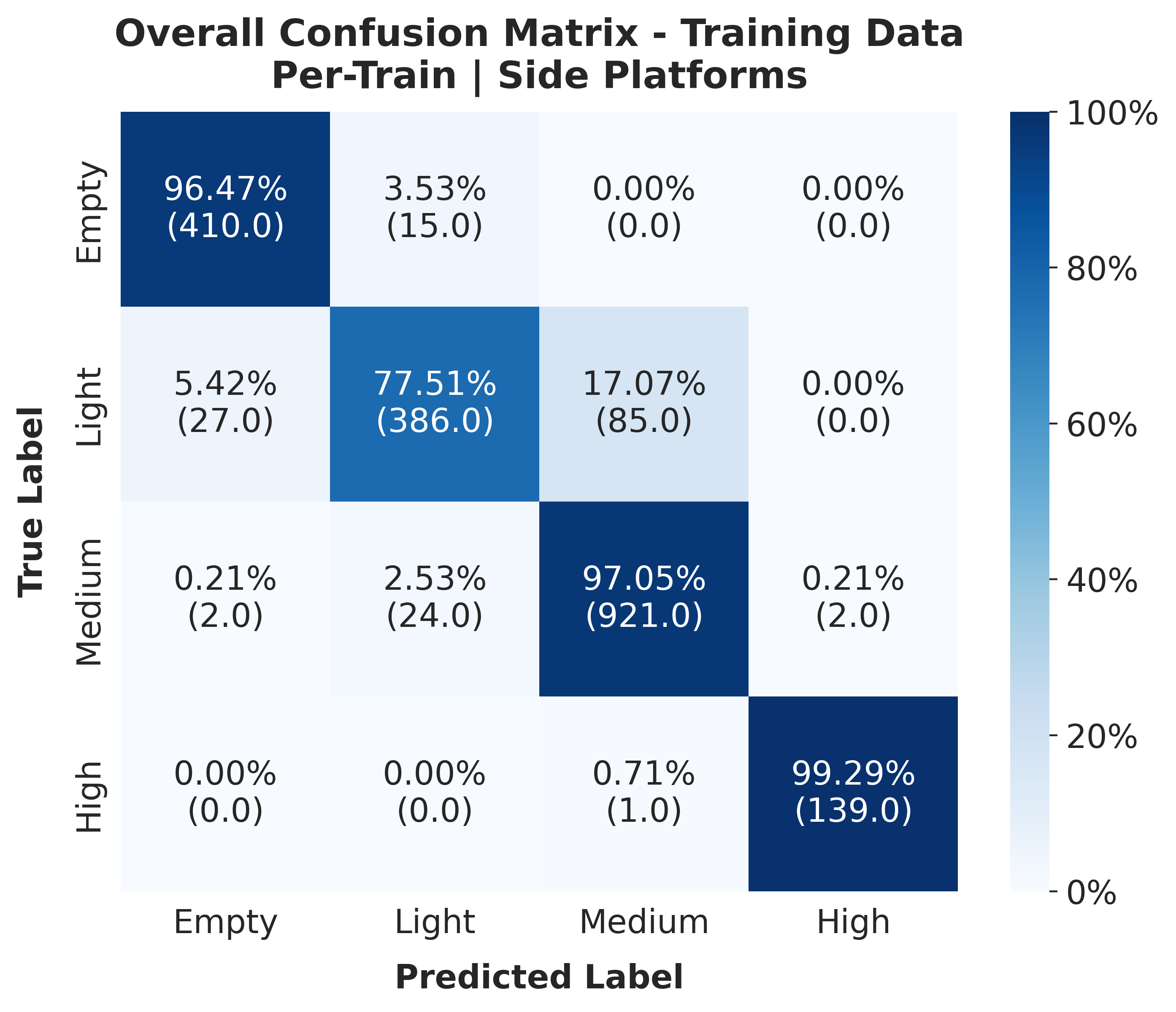}
    }


    \subfigure[Testing set - center platforms.
    \label{fig:conf_matrix_test_center}]{
        \includegraphics[width=0.45\linewidth]{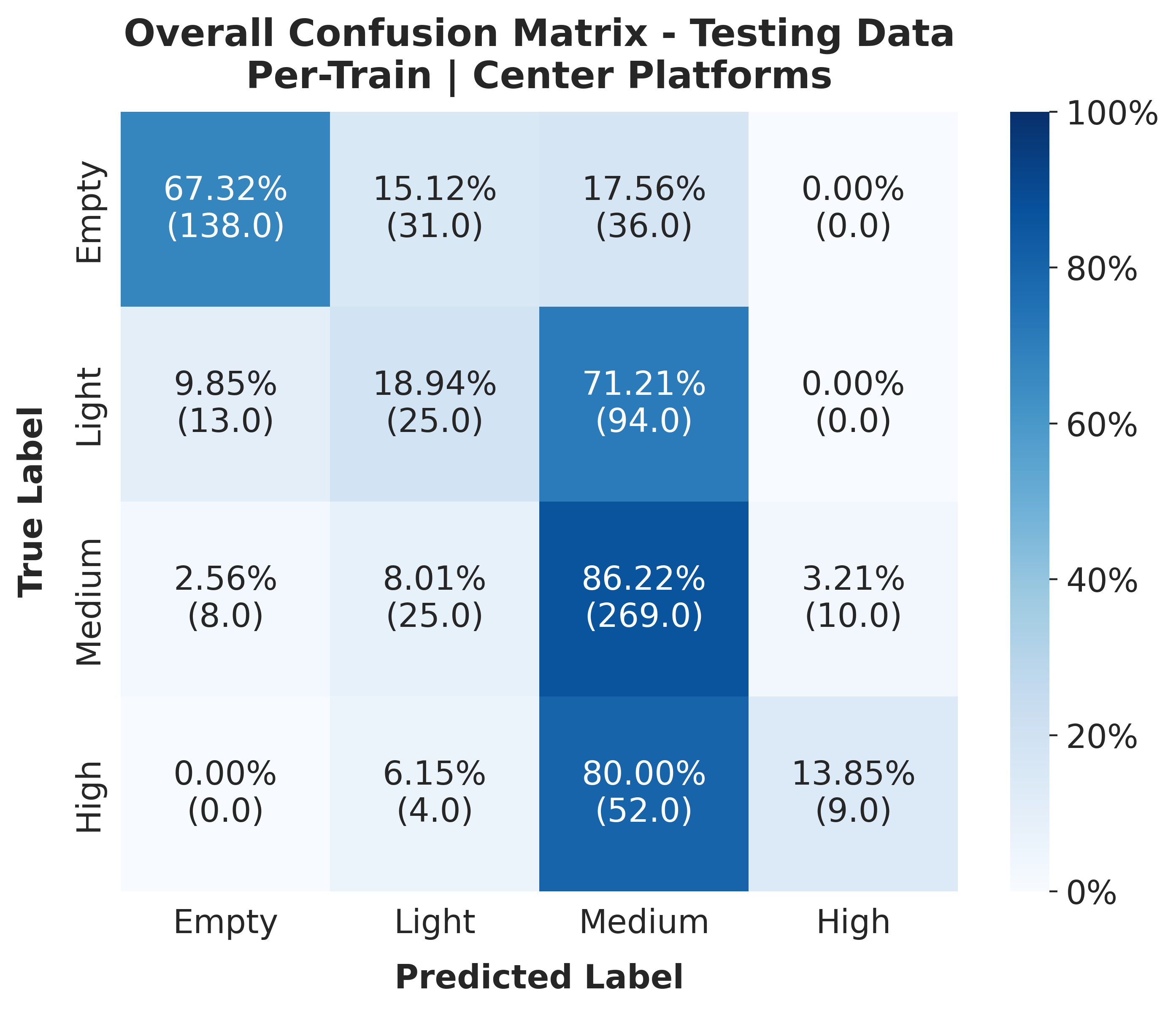}
    }
    \hfill
    \subfigure[Testing set - side platforms.
    \label{fig:conf_matrix_test_side}]{
        \includegraphics[width=0.45\linewidth]{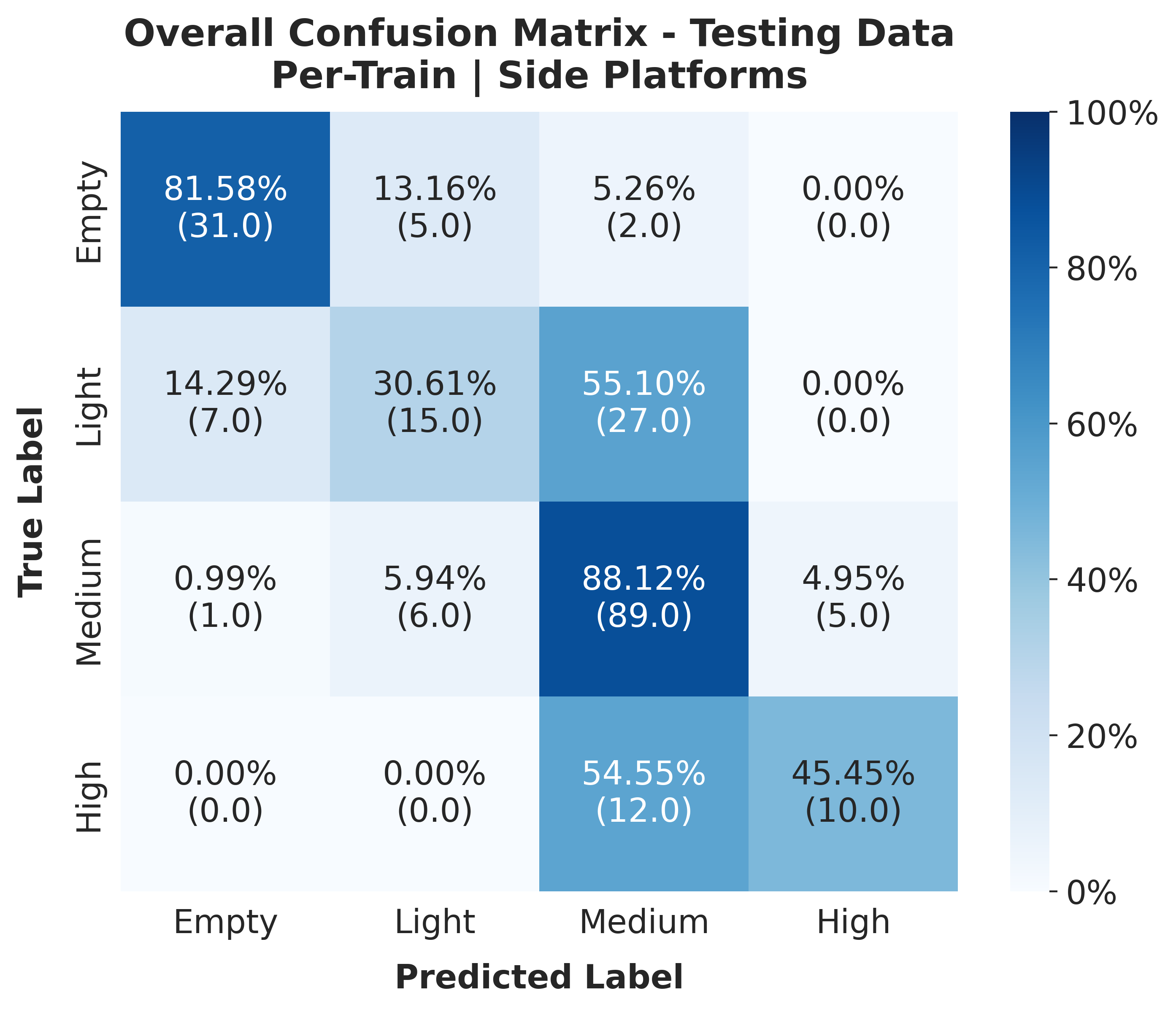}
    }

    \caption{Confusion matrices for per-platform Crowd-ViT models, evaluated on classifying \textit{per-train-arrival event} platform crowd levels, shown by platform type. Values are normalized by row, with diagonals indicating correct classifications.}
    \label{fig:conf_matrices_clustered}
\end{figure}

The performance gap reflects overfitting, likely due to the ViT's 86 million parameters and limited training data. Still, confusion matrices (\autoref{fig:conf_matrices_clustered}) show that misclassifications largely occur between adjacent crowd levels, suggesting the model captures ordinal structure. No extreme misclassifications (e.g., \textit{very light} as \textit{high}) were observed. Here, limitations arising from our relatively small image dataset (compared to what is typically needed for training a vision transformer model), noise from our automatic label-generation process, and the crowd-level boundaries defined by occupancy percentiles are all areas for improvement to reduce the observed misclassifications. Despite modest accuracy, the model still provides useful categorical signals, allowing for a clear distinction between very light and crowded scenes.

\subsection{Segmentation}
\label{ch3:results:segmentation}

\begin{figure}[!ht]
    \centering
    \subfigure[\textit{Per-image:} Each data point corresponds to a single CCTV frame. The higher temporal resolution introduces noise but captures frame-level variation in crowding.
    \label{fig:two_platforms_per_image_segmentation}]{
        \includegraphics[width=\linewidth]{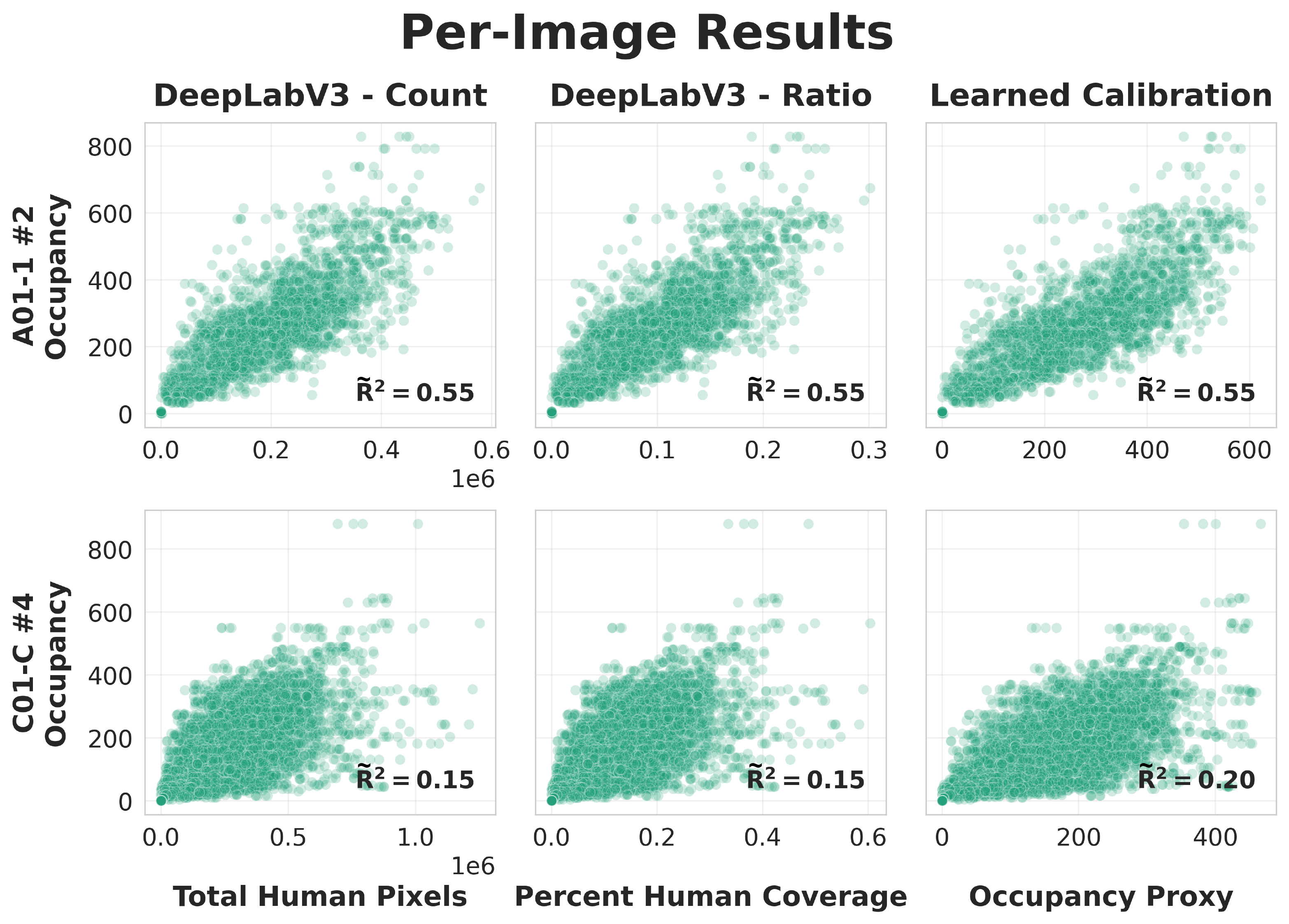}
    }
    \hfill
    \subfigure[\textit{Per-train-arrival event:} Mean segmentation value across all frames per arrival. Temporal aggregation smooths frame-level fluctuations and sharpens the relationship to ODX-derived reference occupancy.
    \label{fig:two_platforms_per_train_segmentation}]{
        \includegraphics[width=\linewidth]{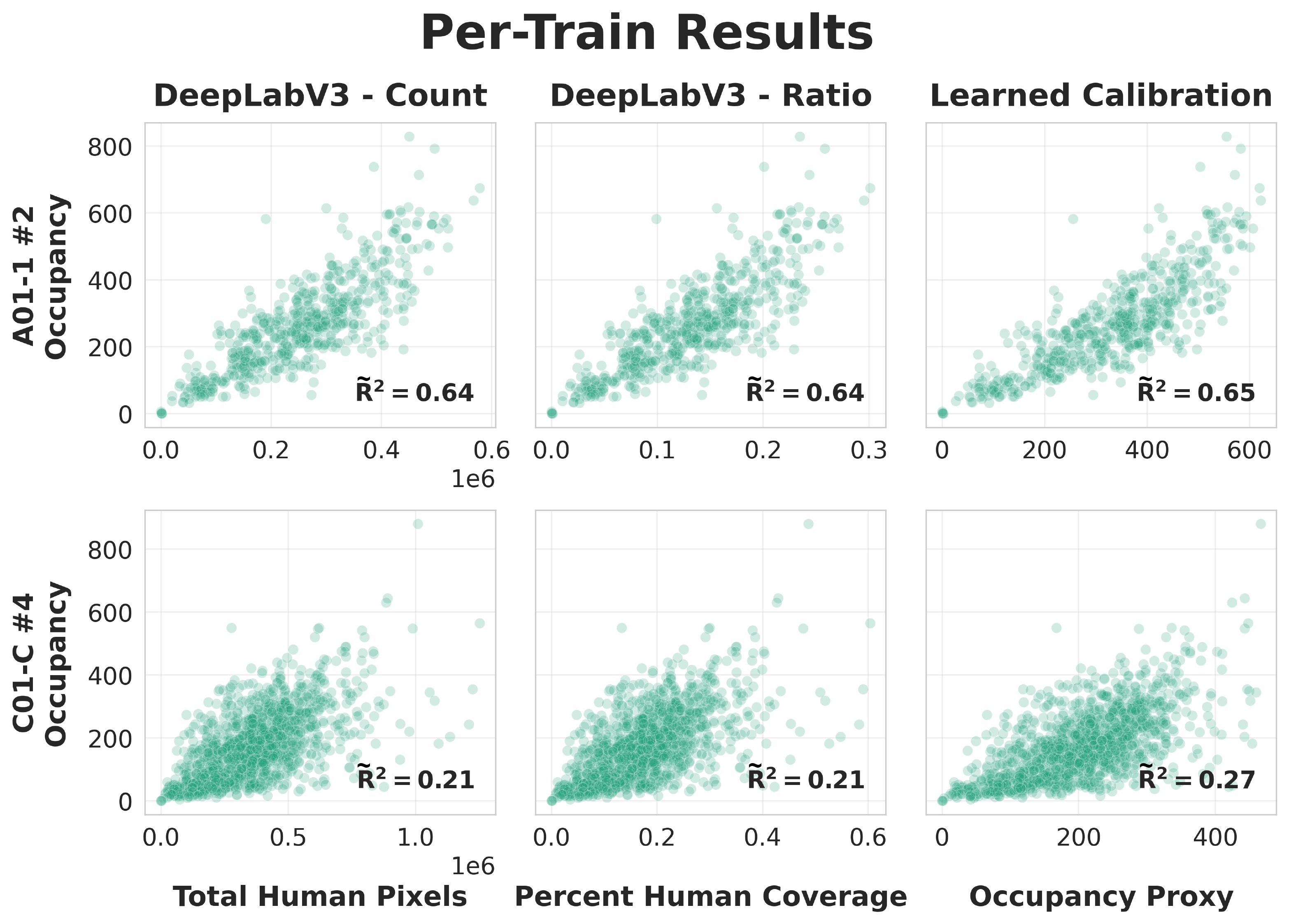}
    }
    \caption{Semantic segmentation (DeepLabV3) features plotted against ODX-derived reference occupancy. In both figures, \textbf{Top:} Metro Center Red line platform A01-2 (side platform). \textbf{Bottom:} Metro Center Blue/Orange/Silver line platform C01-C (center platform).}
    \label{fig:two_platforms_segmentation_models}
\end{figure}

We evaluate segmentation-derived estimates of platform occupancy at per-image and per-train-arrival event levels using the DeepLabV3 semantic segmentation model outputs. As outlined earlier, three features are extracted: (i) raw binary (person/no-person) pixel counts, (ii) the ratio of these pixels within platform-specific regions of interest (see \autoref{fig:mask-overlay}), and (iii) calibrated estimates derived via camera-specific spatial RR calibration, applied only to cameras with at least 600 annotated samples in the training set (platforms A01-1, A01-2, and C01-C).

\autoref{fig:two_platforms_segmentation_models} shows strong correlations among these segmentation-based representations, where calibration improves agreement with ODX-derived reference occupancy. $\tilde{R}^2$ values remain moderate, with errors increasing at mid-to-high occupancy levels, similar to detection-based methods. Side platforms, such as A01-1, demonstrate particularly strong performance due to favorable camera angles, where calibration markedly improves model accuracy. However, pixel-based methods tend to over-count relative to ODX-derived reference occupancy, notably for camera \#2 on platform A01-1.

Certain cameras exhibit non-linear, saturating response patterns, suggesting that fixed-pixel mappings may inadequately capture the spatial complexities of crowds in dense or distorted scenes, even with calibration.

As observed with object detection models, mean-pooling segmentation features across the three frames surrounding each train-arrival event's magic moment reduces variability. It enhances estimation stability (see \autoref{fig:two_platforms_per_train_segmentation}), consistent with findings from object detection methods.

Overall, calibrated segmentation estimates consistently outperform raw pixel-based methods. Nevertheless, uncalibrated ratios and pixel counts remain strong predictors, highlighting the utility of segmentation-based features, particularly in scenarios with limited calibration data.

\subsection{Cross-Camera Per Train-Arrival Event Estimation}
\label{ch3:results:per_train_estimation}

Most platforms are monitored by multiple CCTV cameras, allowing various views on the same train-arrival event to estimate platform crowding. Thus, we leverage this multi-view setup using an LGBM model to integrate per-camera CV-derived features into a unified platform-level occupancy estimate for each train-arrival event. Due to LGBM's robustness to missing values, the model can accommodate partial input, i.e., when not all camera views are available for a given train-arrival event. No additional features are included in the LGBM feature space to show the value of per-arrival and per-platform CV-based estimates for crowding state estimation. The analysis is restricted to platforms A01-1/2 and C01-C due to their camera-level data coverage.

\begin{figure}[h]
        \centering
        \includegraphics[width=\linewidth]{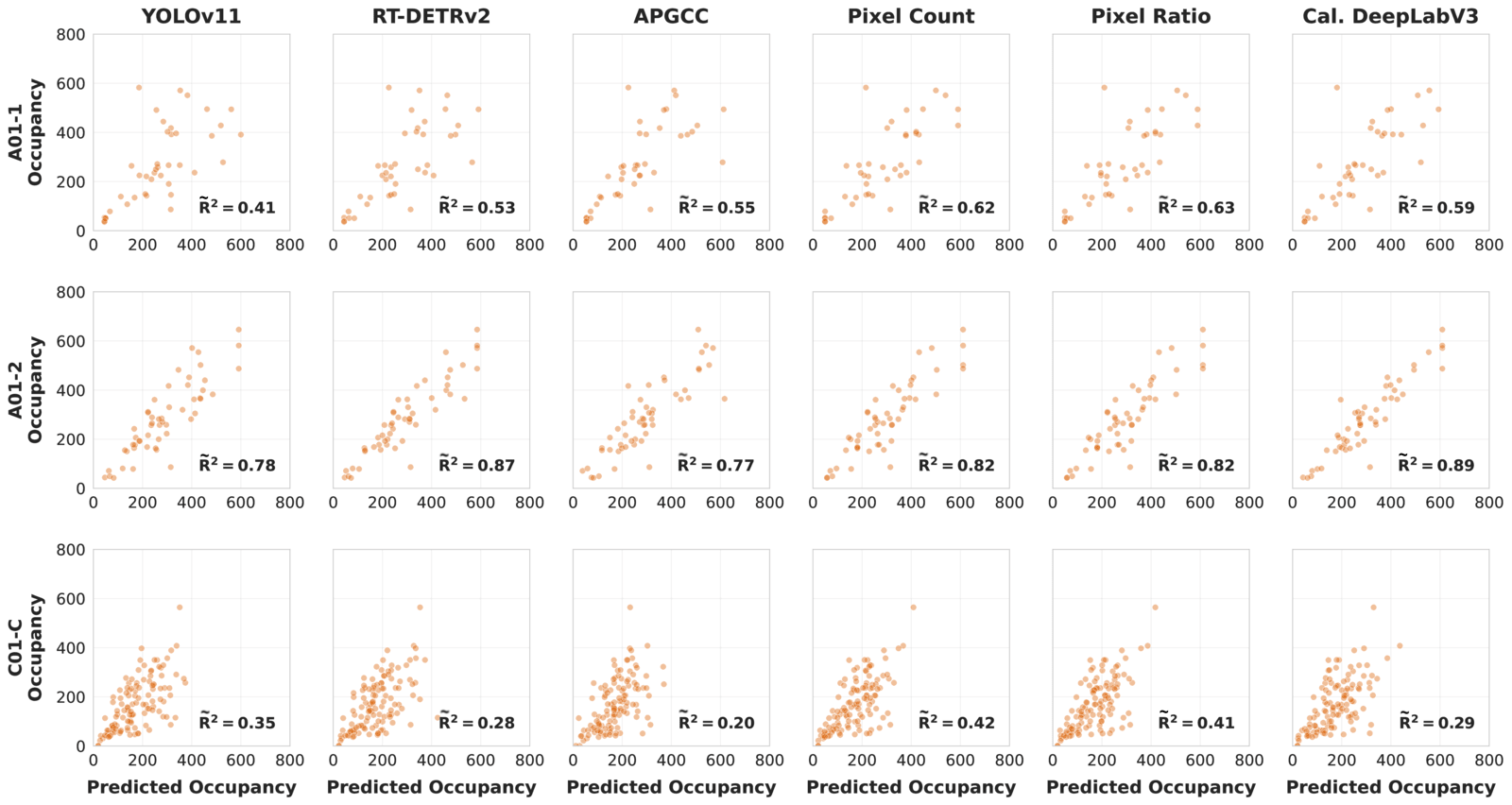}
        \caption{Cross-camera per-train-arrival event platform occupancy estimates using CV-derived features aggregated across multiple cameras. LGBM predictions on the testing dataset use per-train-arrival-event-aggregated CV features across all camera views for platforms A01-1/2 and C01-C.}
        \label{fig:per_train_per_platform_estimates}
\end{figure}

\autoref{fig:per_train_per_platform_estimates} presents the results of such a multi-camera LGBM model, evaluated on the previously stratified testing dataset. While the model successfully learns patterns, substantial variation remains, particularly at high occupancy levels. Performance varies across platforms and CV methods, with no single method universally dominant in terms of $\tilde{R}^2$ values.

\autoref{tab:per_train_arrival} summarizes performance across the three evaluated platforms. The calibrated DeepLabV3 model achieves the strongest overall results, yielding the lowest MAE (56.42), 95\textsuperscript{th} AE (164.51), and wMAE (68.66), together with the highest $\tilde{R}^2$ (0.64). Its weighted error is 2.79\% lower than that of the raw pixel counts from the same segmentation mask and 7.37\% lower than the strongest detection-based model (RT-DETRv2). To examine error patterns across loading conditions, we also report wMAE within the 0\thh-50\thh, 50\thh-75\thh, 75\thh-98\thh, and 98\thh-100\thh$\;$ percentiles. The stratified values generally show the expected increase in error with increasing occupancy, and indicate that the calibrated DeepLabV3 model remains competitive across all regimes, including the 98\thh-100\thh$\;$ one, where errors are largest.

\begin{table}[!ht]
        \centering
        \begin{minipage}[c]{\linewidth}
                \centering
                \caption{Performance of CV-based per-train-arrival event platform occupancy estimates. Results are averaged over platforms A01-1/2 and C01-C. The calibrated DeepLabV3 segmentation approach achieves the lowest absolute errors and the highest standardized agreement ($\tilde{R}^2$). MAE, 95th AE, and wMAE are reported in occupancy counts, while $\tilde{R}^2$ is scale-invariant, with $1$ denoting perfect agreement after scaling.}
                \label{tab:per_train_arrival}
                \resizebox{\linewidth}{!}{%
                \begin{tabular}{lrrrrrrrr}
                        \toprule
                        \textbf{Model} & \textbf{MAE} & \textbf{95\thh$\;$AE} & \textbf{wMAE$\,\downarrow$} & $\mathbf{\tilde{R}^2}$ & wMAE\textsuperscript{0-50} & wMAE\textsuperscript{50-75} & wMAE\textsuperscript{75-98} & wMAE\textsuperscript{98-100} \\
                        \midrule
                        YOLOv11 & 67.87 & 176.41 & 80.18 & 0.57 & 47.33 & 85.55 & 108.05 & 171.96 \\
                        Crowd-ViT & 68.58 & 181.02 & 78.11 & 0.62 & 56.71 & 59.43 & 114.16 & 174.45 \\
                        APGCC & 64.72 & 201.22 & 75.50 & 0.57 & 44.43 & 69.72 & 100.20 & 220.82 \\
                        RT-DETRv2 & 63.06 & 169.00 & 74.12 & 0.61 & 47.73 & 80.97 & 90.20 & 150.44 \\
                        Pixel Count & 59.22 & 168.19 & 70.63 & 0.63 & 48.62 & 62.97 & 92.28 & 153.10 \\
                        	\textbf{Cal. DeepLabV3} & \textbf{56.42} & \textbf{164.51} & \textbf{68.66} & \textbf{0.64} & 51.33 & 50.12 & 90.83 & 142.39 \\
                        \bottomrule
                \end{tabular}%
                }
        \end{minipage}
\end{table}

\subsection{Joint Per-15-minute Bin Estimation}
\label{ch3:joint_ber_bin_estimates}

Building on the per-train-arrival results, we next estimate the mean 15-minute-bin platform occupancy by aggregating the LGBM per-arrival predictions within each bin. This setup evaluates how well the CV-derived features support short-horizon platform occupancy estimation once individual arrivals are pooled. As shown in \autoref{tab:per_15min_bin}, the calibrated DeepLabV3 model again performs best in terms of MAE and wMAE, yielding a 2.15\% wMAE reduction relative to APGCC, although it does not achieve the lowest 95\thh$\;$AE or highest $\tilde{R}^2$.

To examine how performance varies with loading conditions in this coarser temporal setting, we stratify the 15-minute bins by empirical occupancy percentiles. For each platform, the test-set ODX-derived reference occupancies are partitioned into the same percentile ranges as in the preceding section, and wMAE is computed separately for each stratum using the same per-bin error weights. These strata are used solely for diagnostic evaluation and are distinct from the ODX-based crowd-level boundaries applied in the per-train-arrival analysis due to the applied aggregation.

The resulting patterns mirror the per-arrival findings. All CV approaches exhibit increasing variance at higher occupancies, reflecting saturation in dense scenes and reduced visibility. Moreover, averaging across arrivals leaves systematic underestimation at high loads and occasional overestimation at low loads largely intact, since aggregation does not correct model-specific scaling errors. The 15-minute aggregation reduces the number of observations in the 98\thh-100\thh$\;$percentile range (crowd level \textit{high}), making the corresponding error estimates less stable and more sensitive to individual high-error observations. For Crowd-ViT, accuracy is highest in bins dominated by \textit{very light} and \textit{medium} scenes, consistent with the level-imbalance patterns visible in its confusion matrices. RT-DETRv2 attains the lowest 95\textsuperscript{th} AE, reflecting strong performance in low-occupancy regimes and stable mid-range accuracy, although its overall wMAE remains above that of the calibrated DeepLabV3 segmentation model.

\begin{table}[!ht]
    \centering
    \begin{minipage}[c]{\linewidth}
        \centering
        \caption{Results for 15-minute bin mean occupancy estimation computed from multi-camera, per-train-arrival-event LGBM estimates.}
        \label{tab:per_15min_bin}
        \resizebox{\linewidth}{!}{%
        \begin{tabular}{lrrrrrrrr}
            \toprule
            \textbf{Model} & \textbf{MAE} & \textbf{95\thh$\;$AE} & \textbf{wMAE$\,\downarrow$} & $\mathbf{\tilde{R}^2}$ & wMAE\textsuperscript{0-50} & wMAE\textsuperscript{50-75} & wMAE\textsuperscript{75-98} & wMAE\textsuperscript{98-100} \\
            \midrule
            YOLOv11                 & 61.74          & 157.35           & 67.96          & 0.51            & 47.95 & 71.27 & 107.48 & 306.63 \\
            Crowd-ViT               & 61.21          & 153.02           & 64.17          & \textbf{0.61}   & 67.31 & 83.18 & 60.07  & 247.74 \\
            Pixel Count             & 57.35          & 154.71           & 61.79          & 0.57            & 47.59 & 66.98 & 114.04 & 250.73 \\
            RT-DETRv2               & 55.64          & \textbf{144.13}  & 61.44          & 0.58            & 41.62 & 66.43 & 103.73 & 210.04 \\
            APGCC                   & 58.59          & 156.27           & 61.02          & 0.56            & 43.85 & 61.78 & 107.47 & 232.54 \\
            \textbf{Cal. DeepLabV3} & \textbf{54.10} & 152.42           & \textbf{59.71} & 0.57            & 42.20 & 66.96 & 85.49  & 203.19 \\
            \bottomrule
        \end{tabular}%
        }
    \end{minipage}
\end{table}

\subsection{ODX Error Assessment}
\label{seubsec:odx_error_assessment}

\begin{table}[!ht]
    \centering
    \begin{minipage}[c]{\linewidth}
        \centering
        \caption{Estimated latent process noise ($\sigma_u$) and state-space model (SSM)-based ODX observation noise ($\sigma_e$) by platform. Relative values indicate $\sigma_e$ as a percentage of the indicated occupancy statistic.}
        \label{tab:noise_decomposition}
        \resizebox{\linewidth}{!}{%
        \begin{tabular}{lccccccc}
            \toprule
            \textbf{Platform} & $\sigma_u$ & \shortstack{$\sigma_e$} & Median Occ. & Rel.~(Med) & Rel.~(95\thh) & Rel.~(98\thh) \\
            \midrule
            A01-1 & 18.54 & 27.62 & 104.90 & 26.33\% & 7.70\% & 6.10\% \\
            A01-2 & 19.04 & 28.44 & 106.50 & 26.71\% & 8.47\% & 6.82\% \\
            A03-1 & 16.17 & 9.35  & 44.20  & 21.16\% & 6.90\% & 5.56\% \\
            A03-2 & 18.44 & 9.93  & 42.87  & 23.17\% & 7.30\% & 5.47\% \\
            B01-2 & 41.52 & 25.02 & 118.40 & 21.13\% & 8.12\% & 6.65\% \\
            C01-C & 14.81 & 13.64 & 69.83  & 19.54\% & 6.41\% & 5.31\% \\
            F05-C & 20.84 & 20.75 & 32.48  & 63.89\% & 15.60\% & 9.69\% \\
            \bottomrule
        \end{tabular}%
        }
    \end{minipage}
\end{table}

\autoref{tab:noise_decomposition} reports the estimated latent process variability ($\sigma_u$), the SSM-implied ODX observation noise ($\sigma_e$), and the magnitudes of both relative to each platform's median, 95\thh-percentile, and 98\thh-percentile occupancies during 2023.

Across Metro Center, Gallery Place-Chinatown, and DuPont Circle, the SSM-derived ODX measurement noise corresponds to approximately $20$-$30$\% of median occupancy and $6$-$9$\% of 95\thh-98\thh percentile occupancy, computed as $\sigma_e/P_{{50,95,98}}$. Noise estimates are similar across platforms within the same station and higher at Metro Center and Gallery Place-Chinatown, consistent with greater assignment ambiguity at transfer hubs where intra-station transfers are unobserved and concurrent train movements create multiple plausible passenger paths. For the low-occupancy event-oriented station, comparable absolute noise translates into larger relative values.

These relative measures provide a practical reference for evaluating our CCTV-based estimates. Median-based ratios characterize ODX measurement uncertainty under typical conditions, while 95\thh$\;$and 98\thh$\;$percentile ratios quantify uncertainty during peak periods, when accurate crowding estimates are most operationally relevant. Prediction errors of the same order as the reported $\sigma_e$ ratios should therefore be interpreted in light of this uncertainty. As discussed below, however, $\sigma_e$ provides a contextual reference scale for interpreting CV errors and should not be treated as an independently validated measurement-error bound.

\subsubsection{SSM Error Evaluation}
The residuals of our SSM-based reconstruction are approximately Gaussian. However, they exhibit statistically significant autocorrelation, indicating that the simple ARX(1) specification does not capture all temporal structure in the series. This highlights that the various $\sigma_e$ values and the computed ratios should be interpreted as an order-of-magnitude measure of  ODX observation noise modeled under this specification, not as an exact calibration error.

\section{Operational Integration}
\label{ch3:operational_integration}

Integrating CV-based methods, such as object detection, classification, and segmentation, involves computational and operational costs. Empirical estimates of training and inference times, as well as resource requirements from our experiments, are summarized in \autoref{tab:model_timings_requirements}.

\begin{table}[!ht]
    \centering
    \begin{minipage}[c]{\linewidth}
        \centering
        \caption{Per-image resource requirements and timing for single-frame inference, tested on a node with one Nvidia V100 GPU, 20 Intel Xeon-G6 cores, and 170GB RAM.}
        \label{tab:model_timings_requirements}
        \resizebox{\linewidth}{!}{%
        \begin{tabular}{lcccccc}
            \toprule
            & \textbf{YOLOv11 SAHI} & \textbf{RT-DETRv2 SAHI} & \textbf{APGCC} & \textbf{Crowd-ViT}                 & \textbf{DeepLabV3} & \textbf{DeepLabV3 Calibration} \\
            \midrule
            Training [s]     & N/A                   & N/A                     & N/A            & 1.190$\pm$0.068                    & N/A                 & 0.009$\pm$0.007 \\
            Inference [s]    & 3.453$\pm$0.296       & 3.273$\pm$0.852         & 0.25$\pm$0.023 & 0.073$\pm$0.003                    & 0.625$\pm$0.049    & 0.003$\pm$0.001 \\
            Requirements     & CPU/GPU               & CPU/GPU                 & CPU/GPU        & GPU/Data/Occupancy  & CPU/GPU            & CPU/Seg. Maps/Occupancy \\
            \bottomrule
        \end{tabular}%
        }
    \end{minipage}
\end{table}
SAHI-enhanced detection models (YOLOv11, RT-DETRv2) require over 3 seconds per image due to SAHI's multi-patch evaluations (approximately 0.073-0.076 seconds per patch), whereas APGCC, Crowd-ViT, and DeepLabV3 achieve sub-second inference times, suitable for near-real-time applications. In practice, however, inference speed is less restrictive than access to labeled data and platform occupancy counts.

For agencies lacking labeled data, pre-trained detection models such as RT-DETRv2 and APGCC offer readily implementable solutions that balance accuracy and cost-efficiency. RT-DETRv2 can additionally detect object classes such as bicycles and luggage, potentially supporting downstream object-monitoring applications. If threshold-based monitoring is sufficient, segmentation approaches such as DeepLabV3 appeal for their simplicity, speed, and stable outputs. When ODX-derived reference occupancy is available, camera-specific spatial RR calibration allows a direct mapping to occupancy estimates without notably increasing inference times.

We recommend two deployment strategies. For one, agencies with occupancy labels and sufficient computing resources may consider our calibrated segmentation pipeline. In contrast, for those without labeled data, our results suggest using pre-trained detection models, while retaining the option to fine-tune them later through manual labeling. 

Large transformer models (e.g., Crowd-ViT) require substantial labeled data to generalize effectively for classification tasks. Thus, while well-defined operational crowd-level thresholds can improve accuracy, the high data demands limit the advisability of training and employing these models.

In our setting, aggregating multiple frames and camera views has proven to reduce noise, thus improving predictive accuracy. Hence, we recommend capturing multiple frames when estimating crowds to enhance robustness. Furthermore, agencies with a high recurrence of crowding can consider adaptive sampling with lower frequencies during off-peak periods, which could reduce computational requirements.

As noted in the section on \textit{CCTV Privacy Considerations}, WMATA reviewed this study's handling of CCTV imagery in accordance with its internal privacy and data-governance requirements. These requirements should not be interpreted as a general standard for CCTV-based crowd monitoring, since rules governing image retention, storage, access, and model development vary across agencies and jurisdictions. This distinction matters most for models that require agency-specific training data, such as Crowd-ViT, as training and validation can require temporary retention of labeled images, even when all processing remains internal. By contrast, pre-trained models such as YOLOv11, RT-DETRv2, and APGCC can provide usable numeric crowding estimates without task-specific fine-tuning. More broadly, when on-premises or edge computing is available, all evaluated pipelines can be deployed so that raw imagery remains within an agency's environment, eliminating the need for image or footage storage.

\section{Limitations \& Discussion}
\label{ch3:limitations_discussion}

This study demonstrates that CCTV-derived features can support platform occupancy estimation in urban rail environments. Across the evaluated CV methods, calibrated semantic segmentation achieved the strongest overall absolute-error performance, while RT-DETRv2 and Crowd-ViT remained competitive on specific metrics and aggregation levels. Detection and head-counting approaches generally undercounted but benefited from multi-camera aggregation and non-linear rescaling through LGBM.

Performance varied notably between side and center platforms. Side platforms generally exhibited better accuracy, attributed to advantageous camera positions, fewer structural constraints, and no interference of passengers of different trains. Conversely, center platforms, such as Metro Center C01-C, have lower ceilings, limited camera coverage, and greater inter-camera variability, which reduces overall performance. While aggregation methods partially mitigated these issues, further improvements may require further modeling efforts, manual counts to account for rider mixing, or, if possible, optimized camera placements. Although elevated center platforms hinted at better performance, limited data prevented definitive conclusions.

Our first limitation concerns the spatiotemporal coverage of the CCTV dataset, which includes relatively few observations during peak crowding events. This may limit the robustness of the CV- and LGBM-based models under extreme conditions. Expanding data collection to capture a broader range of operational scenarios would strengthen model generalizability.

Another limitation concerns the ODX-derived reference occupancies. As a trip-assignment algorithm, ODX is not designed to provide precise platform occupancy measurements and may introduce noise or misallocation, particularly at transfer stations where intra-station movements are unobserved. Our SSM-based decomposition provides a model-based approximation of this uncertainty, but future work should incorporate limited manual counts or other independent reference measurements.

Despite this limitation, ODX remains the relevant operational benchmark for the present study because the objective is to assess whether CCTV streams can complement existing transit-agency data under real operating conditions, rather than to benchmark CV architectures against manually annotated imagery. Public CV datasets are of limited relevance to this setting because underground transit environments differ substantially in lighting, occlusion, compression, and rider distribution.

Additionally, relying on pre-trained, off-the-shelf models for object detection, head counting, and segmentation without domain-specific fine-tuning limited accuracy. Systematic undercounting was observed in high-density scenes, driven by poor lighting, compression artifacts, occlusions, and variable image quality. Future work should investigate how fine-tuning these models to the transit environment could improve robustness and accuracy. In addition, the camera-specific spatial calibration of the segmentation maps is viewpoint dependent. Thus, if a camera is moved, replaced, zoomed, or substantially reoriented, the learned weight map may require recalibration, which constitutes an operational limitation of the otherwise interpretable calibrated DeepLabV3 approach.

The moderate classification accuracy (63.42\%) of the Crowd-ViT model reflected inherent challenges, including limited support from original samples for rarer crowd levels despite augmentation and visual similarity among neighboring crowd levels. While our findings show that such models can learn to distinguish crowded from non-crowded scenes, enhancing their effectiveness would require notably larger and more balanced datasets, as well as refined and operationally informed crowd-level definitions.

Furthermore, structural constraints inherent in our modeling approach posed challenges. While LGBM effectively handled missing data and provided interpretability, it assumed independent samples, neglecting temporal dependencies and recurrent patterns. Moreover, the LGBM-based model comparisons are restricted to A01-1, A01-2, and C01-C, the three platforms with sufficient multi-camera coverage. Hence, the ranking of calibrated DeepLabV3 relative to the other approaches should not be taken as established network-wide. Integrating real-time visual inputs into advanced temporal models, such as transformers, Long Short-Term Memory (LSTM) networks, or Recurrent Neural Networks (RNNs), combined with traditional transit data sources, could more effectively capture the dynamic nature of crowd movements.

Despite these limitations, image-only CV features achieved strong agreement with ODX-derived occupancies on several platforms, demonstrating their potential for real-time operational use.

The present study estimates platform-level occupancy and does not explicitly resolve passenger movement or localized capacity constraints. A natural extension of the detection and head-counting pipelines would combine temporally dense CCTV sampling with anonymous frame-to-frame association or optical-flow methods, enabling direction-specific counts across predefined virtual lines or zones. Such measures could identify opposing passenger streams and localized accumulations near stairs, escalators, platform access points, or train doors, particularly at transfer stations where bottlenecks may arise despite modest platform-wide occupancy. Because the current dataset samples only three frames around each train-arrival event, these flow measures are not evaluated here and would require dedicated temporal validation and privacy review.

Finally, we situate our findings within the broader literature discussed above. Much of the crowd-counting literature evaluates models on manually annotated, general-purpose benchmark datasets such as CrowdHuman and ShanghaiTech A/B \cite{shao2018crowdhuman, Zhang_2016_CVPR}. In contrast, the closest transit-focused studies have addressed narrower tasks, including image-based passenger counting \cite{goh_image_2018}, in-station crowd-flow estimation and prediction \cite{wong_fusion_2023}, or customized detection on limited platform footage \cite{wang_2025_assessment}. Our setting differs in two key aspects. First, we study fixed platform-mounted CCTV in a real operational rail environment characterized by poor lighting, occlusion, and compression artifacts. Second, we evaluate against ODX-derived operational occupancy estimates rather than manual counts. This enables analysis at greater scale with less manual labeling and demonstrates that CCTV-derived features can support operational platform occupancy estimation under realistic transit conditions.

\section{Conclusion}
\label{ch3:conclusion}

This work explores CCTV imagery as an emerging source for near-real-time platform crowding estimation within the WMATA Metrorail system, leveraging a dataset of 43,232 images collected from seven platforms and 18 cameras. We automatically labeled these images using occupancy estimates from the WMATA-internal integration of the Origin-Destination Interchange (ODX) algorithm, aligned with train-arrival events.

We evaluated various CV approaches, namely, object detection (YOLOv11, RT-DETRv2), head counting (APGCC), crowd-level classification (Crowd-ViT), and calibrated semantic segmentation (DeepLabV3), for their capability to provide platform-level crowding estimates at various temporal resolutions. The DeepLabV3 semantic segmentation model, combined with a camera-specific spatial Ridge Regression (RR) calibration, allows direct occupancy estimation from segmentation features. Furthermore, this approach achieves the strongest overall absolute-error performance among the evaluated approaches on three data-rich platforms.


This research advances platform crowding estimation by demonstrating that CCTV-derived features provide interpretable and operationally relevant insights. Future studies can extend this work by integrating these insights into predictive models for real-time operational decision-making and into time-series forecasting approaches as features.

\begin{acks}
    The authors would like to thank WMATA for supporting this project through its academic research partnership with MIT, and the MIT SuperCloud and Lincoln Laboratory Supercomputing Center for providing high-performance computing resources that have contributed to generating the results reported in this paper. Generative AI provided by Grammarly and OpenAI ChatGPT 4o have been used solely for language editing and has not contributed to study design, data analysis, or the determination of the results and conclusions.
\end{acks}

\begin{funding}
    The authors disclosed the receipt of financial support for the research, authorship, and publication of this article by the Washington Metropolitan Area Transit Authority (WMATA) through its academic research partnership with the Massachusetts Institute of Technology (MIT).
\end{funding}

\begin{contributions}
    The authors confirm their contribution to the paper: Study conception and design: RF, AA, AS, JZ; data collection: RF; analysis and interpretation of results: RF, AA; draft manuscript preparation: RF, AA, AS, JZ, GP, IT. All authors reviewed the results and approved the final version of the manuscript.
\end{contributions}

\begin{dci}
    The authors declared no potential conflicts of interest with respect to the research, authorship, and/or publication of this article.
\end{dci}

\bibliographystyle{SageH}
\bibliography{references}

@article{10462016,
  title    = {Uncertainty quantification of spatiotemporal travel demand with probabilistic graph neural networks},
  volume   = {25},
  issn     = {1558-0016},
  doi      = {10.1109/TITS.2024.3367779},
  number   = {8},
  journal  = {IEEE Transactions on Intelligent Transportation Systems},
  author   = {Wang, Qingyi and Wang, Shenhao and Zhuang, Dingyi and Koutsopoulos, Haris and Zhao, Jinhua},
  month    = aug,
  year     = {2024},
  pages    = {8770--8781}
}

@article{abdelhalim_computer_2024,
  title      = {Computer vision for transit travel time prediction: an end-to-end framework using roadside urban imagery},
  issn       = {1866-749X, 1613-7159},
  shorttitle = {Computer vision for transit travel time prediction},
  url        = {https://link.springer.com/10.1007/s12469-023-00346-3},
  doi        = {10.1007/s12469-023-00346-3},
  language   = {en},
  urldate    = {2024-07-21},
  journal    = {Public Transport},
  author     = {Abdelhalim, Awad and Zhao, Jinhua},
  month      = feb,
  year       = {2024}
}

@article{akyon_slicing_2022,
  title   = {Slicing {Aided} {Hyper} {Inference} and {Fine}-tuning for {Small} {Object} {Detection}},
  url     = {https://ieeexplore.ieee.org/document/9897990},
  doi     = {10.1109/ICIP46576.2022.9897990},
  journal = {2022 IEEE International Conference on Image Processing (ICIP)},
  author  = {Akyon, Fatih Cagatay and Altinuc, Sinan Onur and Temizel, Alptekin},
  year    = {2022},
  pages   = {966--970}
}

@inproceedings{bib:Domahidi2013ecos,
  title     = {{ECOS}: {An} {SOCP} solver for embedded systems},
  booktitle = {European control conference ({ECC})},
  author    = {Domahidi, A. and Chu, E. and Boyd, S.},
  year      = {2013},
  pages     = {3071--3076}
}

@article{chen_deeplab_2017,
  title   = {Deeplab: {Semantic} image segmentation with deep convolutional nets, atrous convolution, and fully connected crfs},
  volume  = {40},
  number  = {4},
  journal = {IEEE transactions on pattern analysis and machine intelligence},
  author  = {Chen, Liang-Chieh and Papandreou, George and Kokkinos, Iasonas and Murphy, Kevin and Yuille, Alan L},
  year    = {2017},
  note    = {Publisher: IEEE},
  pages   = {834--848}
}

@misc{chen_improving_2024,
  title     = {Improving {Point}-based {Crowd} {Counting} and {Localization} {Based} on {Auxiliary} {Point} {Guidance}},
  url       = {http://arxiv.org/abs/2405.10589},
  doi       = {10.48550/arXiv.2405.10589},
  language  = {en},
  urldate   = {2025-03-01},
  publisher = {arXiv},
  author    = {Chen, I.-Hsiang and Chen, Wei-Ting and Liu, Yu-Wei and Yang, Ming-Hsuan and Kuo, Sy-Yen},
  month     = may,
  year      = {2024},
  note      = {arXiv:2405.10589 [cs]}
}

@misc{chen_rethinking_2017,
  title     = {Rethinking {Atrous} {Convolution} for {Semantic} {Image} {Segmentation}},
  url       = {http://arxiv.org/abs/1706.05587},
  doi       = {10.48550/arXiv.1706.05587},
  language  = {en},
  urldate   = {2025-04-15},
  publisher = {arXiv},
  author    = {Chen, Liang-Chieh and Papandreou, George and Schroff, Florian and Adam, Hartwig},
  month     = dec,
  year      = {2017},
  note      = {arXiv:1706.05587 [cs]}
}

@inproceedings{cheng_rethinking_2022,
  title     = {Rethinking {Spatial} {Invariance} of {Convolutional} {Networks} for {Object} {Counting}},
  booktitle = {Proceedings of the {IEEE}/{CVF} {Conference} on {Computer} {Vision} and {Pattern} {Recognition}},
  author    = {Cheng, Zhi-Qi and Dai, Qi and Li, Hong and Song, Jingkuan and Wu, Xiao and Hauptmann, Alexander G.},
  month     = jun,
  year      = {2022},
  pages     = {19638--19648}
}

@article{cvxpy,
  title   = {{CVXPY}: a {Python}-embedded modeling language for convex optimization},
  url     = {https://stanford.edu/~boyd/papers/pdf/cvxpy_paper.pdf},
  journal = {Journal of Machine Learning Research},
  author  = {Diamond, Steven and Boyd, Stephen},
  year    = {2016}
}

@article{dosovitskiy_image_2021,
  title   = {An Image is Worth 16x16 Words: Transformers for Image Recognition at Scale},
  author  = {Alexey Dosovitskiy and Lucas Beyer and Alexander Kolesnikov and Dirk Weissenborn and Xiaohua Zhai and Thomas Unterthiner and Mostafa Dehghani and Matthias Minderer and Georg Heigold and Sylvain Gelly and Jakob Uszkoreit and Neil Houlsby},
  journal = {ArXiv},
  year    = {2020},
  volume  = {abs/2010.11929},
  url     = {https://api.semanticscholar.org/CorpusID:225039882}
}

@book{durbin_time_2012,
  address      = {Oxford},
  edition      = {2nd ed},
  series       = {Oxford {Statistical} {Science} {Ser}},
  title        = {Time {Series} {Analysis} by {State} {Space} {Methods}},
  isbn         = {978-0-19-964117-8 978-0-19-162718-7},
  language     = {en},
  number       = {v.38},
  publisher    = {Oxford University Press, Incorporated},
  author       = {Durbin, James},
  collaborator = {Koopman, Siem Jan},
  year         = {2012}
}

@article{everingham_pascal_2015,
  title      = {The {Pascal} {Visual} {Object} {Classes} {Challenge}: {A} {Retrospective}},
  volume     = {111},
  issn       = {0920-5691, 1573-1405},
  shorttitle = {The {Pascal} {Visual} {Object} {Classes} {Challenge}},
  url        = {http://link.springer.com/10.1007/s11263-014-0733-5},
  doi        = {10.1007/s11263-014-0733-5},
  language   = {en},
  number     = {1},
  urldate    = {2025-04-16},
  journal    = {International Journal of Computer Vision},
  author     = {Everingham, Mark and Eslami, S. M. Ali and Van Gool, Luc and Williams, Christopher K. I. and Winn, John and Zisserman, Andrew},
  month      = jan,
  year       = {2015},
  pages      = {98--136}
}

@inproceedings{florez_monitoring_2023,
  address   = {Cham},
  title     = {Monitoring {System} for {Detecting} {Non}-inclusive {Situations} in {Smart} {Cities}},
  isbn      = {978-3-031-36957-5},
  booktitle = {Trends in {Sustainable} {Smart} {Cities} and {Territories}},
  publisher = {Springer Nature Switzerland},
  author    = {Florez, Sebastian Lopez and Briones, Alfonso González and Pavón, Juan and Fuentes-Fernández, Rubén and Corchado, Juan Manuel},
  editor    = {Castillo Ossa, Luis Fernando and Isaza, Gustavo and Cardona, Oscar and Castrillón, Omar Danilo and Corchado Rodriguez, Juan Manuel and De la Prieta Pintado, Fernando},
  year      = {2023},
  pages     = {405--416}
}

@inproceedings{goh_image_2018,
  title     = {Image {Analytics} for {Train} {Crowd} {Estimation}},
  doi       = {10.1109/DICTA.2018.8615794},
  booktitle = {2018 {Digital} {Image} {Computing}: {Techniques} and {Applications} ({DICTA})},
  author    = {Goh, Choon Giap and Lim, Wee Han and Chua, Justus and Atmosukarto, Indriyati},
  month     = dec,
  year      = {2018},
  pages     = {1--6}
}

@article{gordon_estimation_2018,
  title    = {Estimation of population origin–interchange–destination flows on multimodal transit networks},
  volume   = {90},
  issn     = {0968090X},
  url      = {https://linkinghub.elsevier.com/retrieve/pii/S0968090X18303255},
  doi      = {10.1016/j.trc.2018.03.007},
  language = {en},
  urldate  = {2024-07-31},
  journal  = {Transportation Research Part C: Emerging Technologies},
  author   = {Gordon, Jason B. and Koutsopoulos, Haris N. and Wilson, Nigel H.M.},
  month    = may,
  year     = {2018},
  pages    = {350--365}
}

@misc{jocher_ultralytics_2023,
  title  = {Ultralytics {YOLO}},
  url    = {https://github.com/ultralytics/ultralytics},
  author = {Jocher, Glenn and Chaurasia, Ayush and Qiu, Jing},
  month  = jan,
  year   = {2023}
}

@article{khan_advances_2020,
  title    = {Advances and {Trends} in {Real} {Time} {Visual} {Crowd} {Analysis}},
  volume   = {20},
  issn     = {1424-8220},
  url      = {https://www.mdpi.com/1424-8220/20/18/5073},
  doi      = {10.3390/s20185073},
  number   = {18},
  journal  = {Sensors},
  author   = {Khan, Khalil and Albattah, Waleed and Khan, Rehan Ullah and Qamar, Ali Mustafa and Nayab, Durre},
  year     = {2020}
}

@inproceedings{kopsidas_extracting_2023,
  address   = {Cham},
  title     = {Extracting {Metro} {Passenger} {Flow} {Predictors} from {Network}'s {Complex} {Characteristics}},
  isbn      = {978-3-031-21127-0},
  booktitle = {Complex {Networks} and {Their} {Applications} {XI}},
  publisher = {Springer International Publishing},
  author    = {Kopsidas, Athanasios and Douvaras, Aristeides and Kepaptsoglou, Konstantinos},
  editor    = {Cherifi, Hocine and Mantegna, Rosario Nunzio and Rocha, Luis M. and Cherifi, Chantal and Miccichè, Salvatore},
  year      = {2023},
  pages     = {529--540}
}

@inproceedings{lecun_convolutional_2010,
  address   = {Paris, France},
  title     = {Convolutional networks and applications in vision},
  isbn      = {978-1-4244-5308-5},
  url       = {http://ieeexplore.ieee.org/document/5537907/},
  doi       = {10.1109/ISCAS.2010.5537907},
  urldate   = {2024-07-21},
  booktitle = {Proceedings of 2010 {IEEE} {International} {Symposium} on {Circuits} and {Systems}},
  publisher = {IEEE},
  author    = {LeCun, Yann and Kavukcuoglu, Koray and Farabet, Clement},
  month     = may,
  year      = {2010},
  pages     = {253--256}
}

@article{li_crowding_2011,
  title      = {Crowding and public transport: {A} review of willingness to pay evidence and its relevance in project appraisal},
  volume     = {18},
  copyright  = {https://www.elsevier.com/tdm/userlicense/1.0/},
  issn       = {0967070X},
  shorttitle = {Crowding and public transport},
  url        = {https://linkinghub.elsevier.com/retrieve/pii/S0967070X11000849},
  doi        = {10.1016/j.tranpol.2011.06.003},
  language   = {en},
  number     = {6},
  urldate    = {2024-12-03},
  journal    = {Transport Policy},
  author     = {Li, Zheng and Hensher, David A.},
  month      = nov,
  year       = {2011},
  pages      = {880--887}
}

@misc{lin_microsoft_2014,
  title      = {Microsoft {COCO}: {Common} {Objects} in {Context}},
  copyright  = {arXiv.org perpetual, non-exclusive license},
  shorttitle = {Microsoft {COCO}},
  url        = {https://arxiv.org/abs/1405.0312},
  doi        = {10.48550/ARXIV.1405.0312},
  urldate    = {2024-07-29},
  publisher  = {arXiv},
  author     = {Lin, Tsung-Yi and Maire, Michael and Belongie, Serge and Bourdev, Lubomir and Girshick, Ross and Hays, James and Perona, Pietro and Ramanan, Deva and Zitnick, C. Lawrence and Dollár, Piotr},
  year       = {2014},
  note       = {Version Number: 3}
}

@misc{lv_rt-detrv2_2024,
  title      = {{RT}-{DETRv2}: {Improved} {Baseline} with {Bag}-of-{Freebies} for {Real}-{Time} {Detection} {Transformer}},
  shorttitle = {{RT}-{DETRv2}},
  url        = {http://arxiv.org/abs/2407.17140},
  doi        = {10.48550/arXiv.2407.17140},
  language   = {en},
  urldate    = {2025-04-15},
  publisher  = {arXiv},
  author     = {Lv, Wenyu and Zhao, Yian and Chang, Qinyao and Huang, Kui and Wang, Guanzhong and Liu, Yi},
  month      = jul,
  year       = {2024},
  note       = {arXiv:2407.17140 [cs]}
}

@article{miller_estimation_2018,
  title    = {Estimation of {Passengers} {Left} {Behind} by {Trains} in {High}-{Frequency} {Transit} {Service} {Operating} {Near} {Capacity}},
  volume   = {2672},
  issn     = {0361-1981, 2169-4052},
  url      = {http://journals.sagepub.com/doi/10.1177/0361198118794291},
  doi      = {10.1177/0361198118794291},
  language = {en},
  number   = {8},
  urldate  = {2024-07-22},
  journal  = {Transportation Research Record: Journal of the Transportation Research Board},
  author   = {Miller, Eli and Sánchez-Martínez, Gabriel E. and Nassir, Neema},
  month    = dec,
  year     = {2018},
  pages    = {497--504}
}

@inproceedings{NIPS2017_6449f44a,
  title     = {{LightGBM}: a highly efficient gradient boosting decision tree},
  volume    = {30},
  url       = {https://proceedings.neurips.cc/paper_files/paper/2017/file/6449f44a102fde848669bdd9eb6b76fa-Paper.pdf},
  booktitle = {Advances in neural information processing systems},
  publisher = {Curran Associates, Inc.},
  author    = {Ke, Guolin and Meng, Qi and Finley, Thomas and Wang, Taifeng and Chen, Wei and Ma, Weidong and Ye, Qiwei and Liu, Tie-Yan},
  editor    = {Guyon, I. and Luxburg, U. Von and Bengio, S. and Wallach, H. and Fergus, R. and Vishwanathan, S. and Garnett, R.},
  year      = {2017},
  pages     = {3146--3154}
}

@inproceedings{NIPS2017_8a20a862,
  title     = {A unified approach to interpreting model predictions},
  volume    = {30},
  url       = {https://proceedings.neurips.cc/paper_files/paper/2017/file/8a20a8621978632d76c43dfd28b67767-Paper.pdf},
  booktitle = {Advances in neural information processing systems},
  publisher = {Curran Associates, Inc.},
  author    = {Lundberg, Scott M and Lee, Su-In},
  editor    = {Guyon, I. and Luxburg, U. Von and Bengio, S. and Wallach, H. and Fergus, R. and Vishwanathan, S. and Garnett, R.},
  year      = {2017},
  pages     = {4765--4774}
}

@article{noursalehi_predictive_2021,
  title    = {Predictive decision support platform and its application in crowding prediction and passenger information generation},
  volume   = {129},
  issn     = {0968090X},
  url      = {https://linkinghub.elsevier.com/retrieve/pii/S0968090X21001571},
  doi      = {10.1016/j.trc.2021.103139},
  language = {en},
  urldate  = {2024-03-30},
  journal  = {Transportation Research Part C: Emerging Technologies},
  author   = {Noursalehi, Peyman and Koutsopoulos, Haris N. and Zhao, Jinhua},
  month    = aug,
  year     = {2021},
  pages    = {103139}
}

@misc{pekingu_rtdetr,
  author = {{Peking University}},
  title  = {{RT-DETRv2 R101VD}},
  year   = {2024},
  url    = {https://huggingface.co/PekingU/rtdetr_v2_r101vd},
  note   = {Hugging Face model repository}
}

@article{peppa_urban_2018,
  title     = {Urban {Traffic} {Flow} {Analysis} {Based} on {Deep} {Learning} {Car} {Detection} from {CCTV} {Image} {Series}},
  volume    = {XLII-4},
  copyright = {https://creativecommons.org/licenses/by/4.0/},
  issn      = {2194-9034},
  url       = {https://isprs-archives.copernicus.org/articles/XLII-4/499/2018/},
  doi       = {10.5194/isprs-archives-XLII-4-499-2018},
  language  = {en},
  urldate   = {2024-07-21},
  journal   = {The International Archives of the Photogrammetry, Remote Sensing and Spatial Information Sciences},
  author    = {Peppa, M. V. and Bell, D. and Komar, T. and Xiao, W.},
  month     = sep,
  year      = {2018},
  pages     = {499--506}
}

@inproceedings{Redmon_2016_CVPR,
  title     = {You only look once: {Unified}, real-time object detection},
  booktitle = {Proceedings of the {IEEE} conference on computer vision and pattern recognition ({CVPR})},
  author    = {Redmon, Joseph and Divvala, Santosh and Girshick, Ross and Farhadi, Ali},
  month     = jun,
  year      = {2016},
  pages     = {779--788}
}

@inproceedings{ronneberger2015u,
  title     = {U-net: {Convolutional} networks for biomedical image segmentation},
  booktitle = {Medical image computing and computer-assisted intervention–{MICCAI} 2015: 18th international conference, {Munich}, {Germany}, {October} 5-9, 2015, proceedings, part {III} 18},
  publisher = {Springer},
  author    = {Ronneberger, Olaf and Fischer, Philipp and Brox, Thomas},
  year      = {2015},
  pages     = {234--241}
}

@article{saleh_recent_2015,
  title    = {Recent survey on crowd density estimation and counting for visual surveillance},
  volume   = {41},
  issn     = {0952-1976},
  url      = {https://www.sciencedirect.com/science/article/pii/S0952197615000081},
  doi      = {https://doi.org/10.1016/j.engappai.2015.01.007},
  journal  = {Engineering Applications of Artificial Intelligence},
  author   = {Saleh, Sami Abdulla Mohsen and Suandi, Shahrel Azmin and Ibrahim, Haidi},
  year     = {2015},
  pages    = {103--114}
}

@unpublished{shao2018crowdhuman,
  title      = {Crowdhuman: {{A}} Benchmark for Detecting Human in a Crowd},
  author     = {Shao, Shuai and Zhao, Zijian and Li, Boxun and Xiao, Tete and Yu, Gang and Zhang, Xiangyu and Sun, Jian},
  date       = {2018},
  year       = {2018},
  eprint     = {1805.00123},
  eprinttype = {arXiv},
  note       = {arXiv:1805.00123 [cs.CV]}
}

@article{tirachini_crowding_2013,
  title      = {Crowding in public transport systems: {Effects} on users, operation and implications for the estimation of demand},
  volume     = {53},
  issn       = {09658564},
  shorttitle = {Crowding in public transport systems},
  url        = {https://linkinghub.elsevier.com/retrieve/pii/S0965856413001146},
  doi        = {10.1016/j.tra.2013.06.005},
  language   = {en},
  urldate    = {2024-07-30},
  journal    = {Transportation Research Part A: Policy and Practice},
  author     = {Tirachini, Alejandro and Hensher, David A. and Rose, John M.},
  month      = jul,
  year       = {2013},
  pages      = {36--52}
}

@misc{ultralytics_repo,
  title  = {Ultralytics {GitHub} repository},
  url    = {https://github.com/ultralytics/ultralytics?tab=readme-ov-file},
  author = {{Ultralytics}},
  year   = {2025}
}

@misc{vainf_deeplabv3,
  title  = {{DeepLabV3Plus}-{PyTorch}},
  url    = {https://github.com/VainF/DeepLabV3Plus-Pytorch},
  author = {{VainF}},
  year   = {2025}
}

@misc{wang_2025_assessment,
  title     = {Assessment of passenger flow on metro platform based on video images},
  language  = {en},
  publisher = {Beijing Jiaotong University; CRSC Research \& Design Institute Group Co., Ltd.; Beijing Mass Transit Railway Operation Corp., Ltd.; Beijing Metro Operation Co., Ltd.},
  author    = {Wang, Yuhui and Li, Chong and Dou, Fei and Dang, Jiayi and Wang, Yifei},
  month     = jan,
  year      = {2025},
  note      = {Type: Conference Poster}
}

@article{wong_fusion_2023,
  title    = {Fusion of {CCTV} {Video} and {Spatial} {Information} for {Automated} {Crowd} {Congestion} {Monitoring} in {Public} {Urban} {Spaces}},
  volume   = {16},
  issn     = {1999-4893},
  url      = {https://www.mdpi.com/1999-4893/16/3/154},
  doi      = {10.3390/a16030154},
  language = {en},
  number   = {3},
  urldate  = {2023-09-18},
  journal  = {Algorithms},
  author   = {Wong, Vivian W. H. and Law, Kincho H.},
  month    = mar,
  year     = {2023},
  pages    = {154}
}

@inproceedings{Zhang_2016_CVPR,
  title     = {Single-image crowd counting via multi-column convolutional neural network},
  booktitle = {Proceedings of the {IEEE} conference on computer vision and pattern recognition ({CVPR})},
  author    = {Zhang, Yingying and Zhou, Desen and Chen, Siqin and Gao, Shenghua and Ma, Yi},
  month     = jun,
  year      = {2016},
  pages     = {589--597}
}

@inproceedings{Zhao_2024_CVPR,
  title     = {{DETRs} beat yolos on real-time object detection},
  booktitle = {Proceedings of the {IEEE}/{CVF} conference on computer vision and pattern recognition ({CVPR})},
  author    = {Zhao, Yian and Lv, Wenyu and Xu, Shangliang and Wei, Jinman and Wang, Guanzhong and Dang, Qingqing and Liu, Yi and Chen, Jie},
  month     = jun,
  year      = {2024},
  pages     = {16965--16974}
}

\end{document}